\documentclass{article}

\usepackage{microtype}
\usepackage{graphicx}
\usepackage{subcaption}
\usepackage{booktabs} % for professional tables
\usepackage{multirow}
\usepackage{makecell}

\usepackage[utf8]{inputenc}
\usepackage[table]{xcolor} % Required for \rowcolor
\usepackage{xcolor}
\usepackage{enumitem}
\usepackage{dsfont}

\usepackage{hyperref}

\usepackage[preprint]{icml2026}

\usepackage{amsmath}
\usepackage{amssymb}
\usepackage{mathtools}
\usepackage{amsthm}
\usepackage{array}

\usepackage{tcolorbox}
\tcbuselibrary{skins, breakable}
\definecolor{bg_gray}{RGB}{248, 248, 248}
\definecolor{border_gray}{RGB}{180, 180, 180}
\definecolor{code_blue}{RGB}{0, 50, 200}
\definecolor{example_bg}{RGB}{255, 255, 255}

\definecolor{citeblue}{RGB}{0, 110, 191}
\definecolor{linkred}{RGB}{181, 23, 0}
\hypersetup{
    colorlinks=true,
    citecolor=citeblue,
    linkcolor=linkred,
    urlcolor=citeblue
}

\usepackage[capitalize,noabbrev]{cleveref}

\theoremstyle{plain}

\theoremstyle{definition}

\theoremstyle{remark}

\usepackage[textsize=tiny]{todonotes}

\icmltitlerunning{VideoSEG-O3: A Multi-turn Reinforcement Learning Framework for RVOS}

\begin{document}

\twocolumn[
  % \icmltitle{Temporal-Spatial Chain of Thought: A Multi-turn Reinforcement Learning Framework for Reasoning Video Object Segmentation}
  \icmltitle{VideoSEG-O3: A Multi-turn Reinforcement Learning Framework for Reasoning Video Object Segmentation}

  % It is OKAY to include author information, even for blind submissions: the
  % style file will automatically remove it for you unless you've provided
  % the [accepted] option to the icml2026 package.

  % List of affiliations: The first argument should be a (short) identifier you
  % will use later to specify author affiliations Academic affiliations
  % should list Department, University, City, Region, Country Industry
  % affiliations should list Company, City, Region, Country

  % You can specify symbols, otherwise they are numbered in order. Ideally, you
  % should not use this facility. Affiliations will be numbered in order of
  % appearance and this is the preferred way.
  \icmlsetsymbol{equal}{*}

  \begin{icmlauthorlist}
    \icmlauthor{Ming Dai}{1}
    \icmlauthor{Sen Yang}{2}
    \icmlauthor{Boqiang Duan}{2}
    \icmlauthor{Boyuan Tong}{2}
    \icmlauthor{Jiedong Zhuang}{3}
    \icmlauthor{Wankou Yang}{1}
    \icmlauthor{Jingdong Wang}{2}
  \end{icmlauthorlist}

  \icmlaffiliation{1}{School of Automation, Southeast University, Nanjing, China}
  \icmlaffiliation{2}{Baidu Inc, Beijing, China}
  \icmlaffiliation{3}{College of Information Science \& Electronic Engineering, Zhejiang University, Hangzhou, China}

  \icmlcorrespondingauthor{Wankou Yang}{wkyang@seu.edu.cn}
%   \icmlcorrespondingauthor{Firstname2 Lastname2}{first2.last2@www.uk}

  % You may provide any keywords that you find helpful for describing your
  % paper; these are used to populate the "keywords" metadata in the PDF but
  % will not be shown in the document
  \icmlkeywords{Reasoning Video Object Segmentation, Reinforcement Learning, Multimodal Large Language Models}

  \vskip 0.3in
]

% this must go after the closing bracket ] following \twocolumn[ ...

% This command actually creates the footnote in the first column listing the
% affiliations and the copyright notice. The command takes one argument, which
% is text to display at the start of the footnote. The \icmlEqualContribution
% command is standard text for equal contribution. Remove it (just {}) if you
% do not need this facility.

% Use ONE of the following lines. DO NOT remove the command.
% If you have no special notice, KEEP empty braces:
\printAffiliationsAndNotice{}  % no special notice (required even if empty)
% Or, if applicable, use the standard equal contribution text:
% \printAffiliationsAndNotice{\icmlEqualContribution}

\begin{abstract}
Reasoning Video Object Segmentation (RVOS) demands a sophisticated integration of temporal dynamics, spatial details, and linguistic reasoning to achieve precise pixel-level localization. Existing methods are limited to reasoning over fixed initial inputs and lack the capacity to actively acquire further visual evidence, which is often essential for resolving complex references in long or intricate videos. To address this, we propose \textbf{VideoSEG-O3}, the first multi-turn reinforcement learning framework for RVOS that emulates the human \textit{``coarse-to-fine''} cognitive process. It employs a \textit{multi-turn temporal-spatial chain-of-thought} to capture fine-grained details by iteratively pinpointing critical intervals and keyframes. Additionally, to enable the policy to perceive segmentation quality beyond mere text probability of \texttt{[SEG]} during the RL stage, we introduce \textit{SEG-aware logit calibration}, which integrates pixel-wise segmentation feedback directly into the token-level logits. Furthermore, we design a \textit{decoupled thinking trace} to hierarchically decompose the reasoning process into temporal, spatial, and linguistic dimensions, and construct \textbf{VTS-CoT}, a specialized cold-start dataset featuring comprehensive reasoning trajectories. The code and models will be released at \url{https://github.com/Dmmm1997/VideoSEG-O3}.
% Extensive experiments demonstrate that VideoSEG-O3 achieves advanced performance across 8 mainstream RVOS benchmarks, particularly excelling in long-horizon and complex reasoning tasks.
\end{abstract}

\begin{figure}[t]
    \centering
    \includegraphics[width=0.46\textwidth]{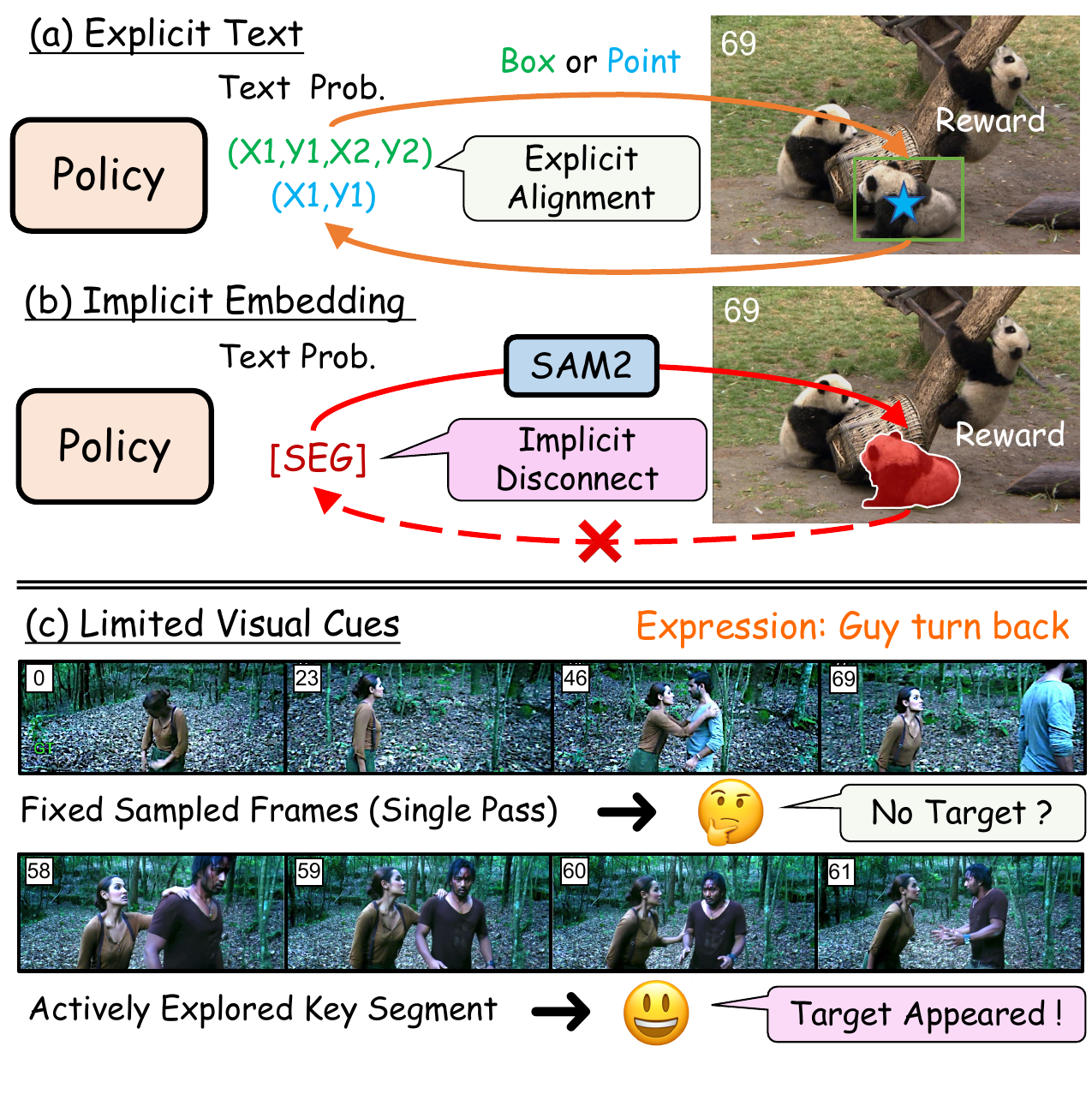} 
    \vspace{-12pt}
    \caption{Motivation of \textbf{VideoSEG-O3}. (a) \textbf{Explicit Text}: Coordinate-based RL methods (e.g., box, point)  allow rewards to directly optimize the policy. (b) \textbf{Implicit Embedding}: Latent mask representations (e.g., \texttt{[SEG]}) suffer from implicit disconnection between token probability and mask quality. (c) \textbf{Limited Visual Cues}: Fixed sampling strategies fail to actively explore key temporal segments required for precise localization.}
    \label{fig:motivation}
    \vspace{-5pt}
\end{figure}

\section{Introduction}

\begin{figure*}[t]
    \centering
    \includegraphics[width=0.95\textwidth]{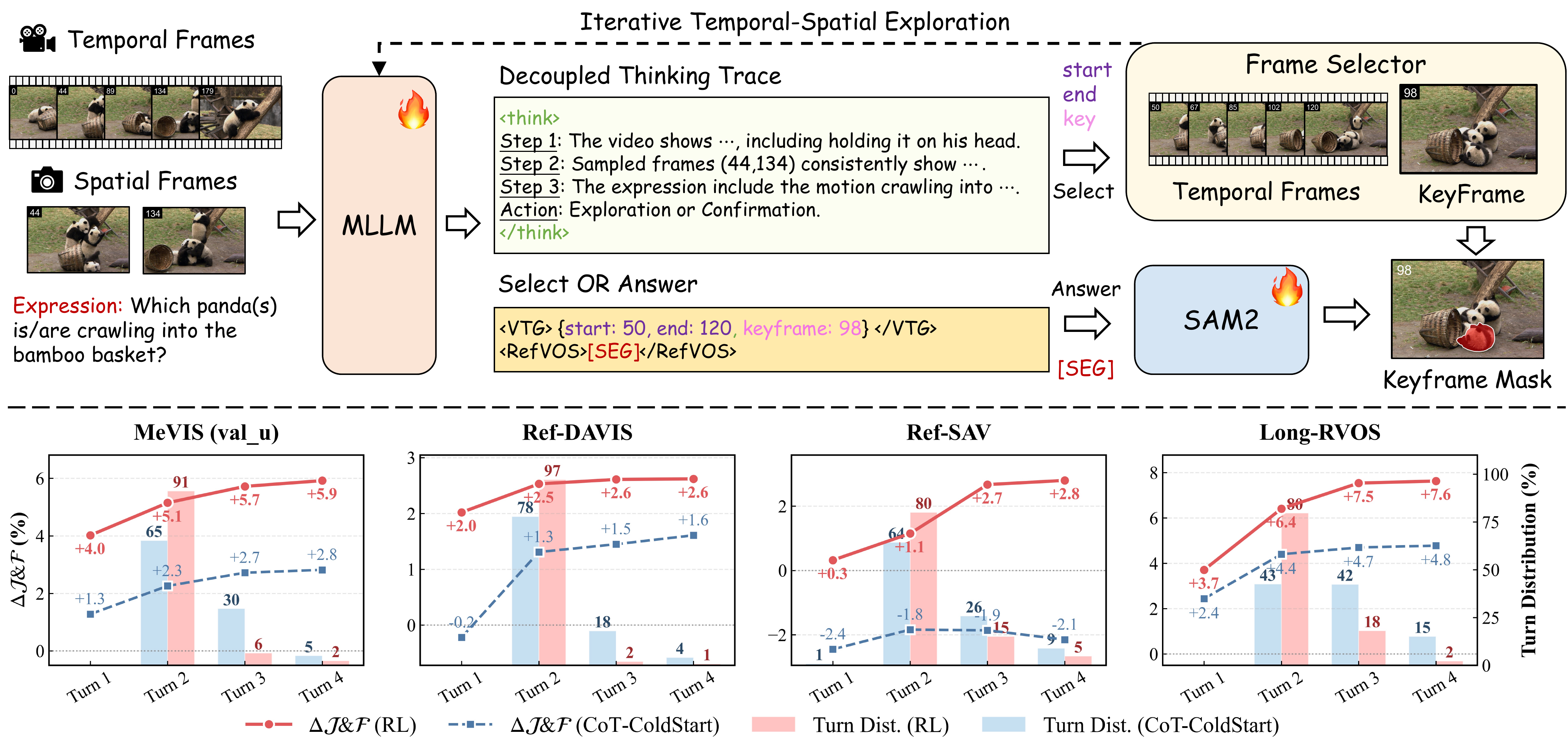} 
    \caption{Overview and statistical analysis of \textbf{VideoSEG-O3}. 
    \textbf{Top:} The MLLM architecture employs a \textit{Decoupled Thinking Trace} to analyze motion context, spatial details, and expressions. A specialized \texttt{<select>} token is utilized to orchestrate iterative temporal-spatial exploration. 
    \textbf{Bottom:} Performance gains ($\Delta \mathcal{J}\&\mathcal{F}$) and turn distributions across four representative benchmarks. Results highlight the performance evolution and adaptive reasoning behavior for both CoT cold-start and RL stages.}
    \label{fig:overall_motivation}
    % \vspace{-5pt}
\end{figure*}

% 现有方法分类
Reasoning Video Object Segmentation (RVOS) aims to localize and segment dynamic objects across video sequences based on complex natural language descriptions~\cite{visa,videolisa,groundmore}. Recently, Multimodal Large Language Models (MLLMs)~\cite{Qwen3-VL, li2024survey} have catalyzed progress in this field by leveraging their robust visual-language understanding capabilities. Existing MLLM-based solutions generally follow two trajectories: \textit{implicit approaches}~\cite{lisa,sa2va,vrt}, which utilize latent embeddings (e.g., \texttt{[SEG]}) to guide mask decoders such as SAM~\cite{sam2}, and \textit{explicit approaches}~\cite{segzero,sam-r1,coprs,lens,omni-r1,Veason_R1}, which generate geometric prompts like bounding boxes or points to trigger external segmenters~\cite{sam2,xmem}.

Within the scope of RL applications, the majority of works~\cite{lens,Veason_R1} favor the explicit paradigm (Fig.~\ref{fig:motivation}(a)). This preference stems from two primary factors: \textit{First}, MLLMs~\cite{Qwen3-VL} possess an inherent proficiency in spatial grounding. \textit{Second}, since geometric prompts are generated as text tokens, the policy can be optimized directly via token-level probabilities. However, the explicit paradigm faces some limitations. \textit{First}, it is inherently indirect, as sparse geometric prompts lack the capacity to accurately characterize fine-grained instance. \textit{Second}, boxes and points are frame-specific, making them insufficient to serve as stable, unified instance representations across a video. In contrast, implicit embeddings can provide consistent descriptors that maintain robust object identity throughout the sequence~\cite{videolisa}.

Furthermore, existing RVOS methods~\cite{liu2025unipixel,onethinker} are restricted to reasoning over a fixed set of sampled frames, lacking the agency to actively seek supplementary visual evidence when the initial context is insufficient to resolve intricate references. As illustrated in Fig.~\ref{fig:motivation}(c), such fixed sampling strategies~\cite{sa2va} inevitably overlook critical context, leading to localization failures. Moreover, these methods fail to utilize visual tokens efficiently. Specifically, conditioned on a specific textual query, the vast majority of visual tokens in a video are often irrelevant. Consequently, processing high-resolution inputs incurs substantial computational redundancy due to the calculation over these uninformative frame tokens.

To address these challenges, we propose \textbf{VideoSEG-O3}, which is governed by three pivotal imperatives: \textbf{(1)} Transitioning from static single-pass inference to a \textit{multi-turn temporal-spatial Chain-of-Thought}. \textbf{(2)} Bridging the modality gap between discrete text generation and dense mask prediction by enabling \textit{direct token-level optimization} for implicit latent embeddings (e.g., \texttt{[SEG]}); and \textbf{(3)} \textit{Decoupling the thinking process} into structured dimensions of temporal dynamics, spatial details, and linguistic expressions to facilitate a more granular understanding.

\noindent\textbf{(1) Multi-turn Temporal-Spatial Chain-of-Thought.} Diverging from single-pass segmentation frameworks, VideoSEG-O3 adopts an active spatiotemporal exploration architecture.
As shown in Fig.~\ref{fig:overall_motivation} (\textbf{Top}), VideoSEG-O3 utilizes the \texttt{<select>} token to orchestrate iterative exploration, dynamically integrating temporal intervals and spatial keyframes. This explicitly decouples temporal action understanding from fine-grained spatial localization. Crucially, the model autonomously determines the optimal reasoning depth, continuously gathering supplementary visual cues to refine the segmentation.
Quantitative analysis in Fig.~\ref{fig:overall_motivation} (\textbf{Bottom}) further validates this design:
\textbf{(1)} \textit{Progressive Gain:} Performance consistently scales with interaction turns across both cold-start and RL stages, demonstrating the effectiveness of additional reasoning processes and the incorporation of supplementary visual cues.
\textbf{(2)} \textit{Reasoning Efficiency:} Post-RL, the model achieves superior performance with reduced interaction turns. This indicates that RL optimizes the select policy to precisely target decisive temporal frames.
\textbf{(3)} \textit{Adaptive Termination:} The model dynamically adjusts its search depth relative to task complexity. It converges in fewer turns on simple benchmarks (e.g., Ref-DAVIS) while allocating additional visual cues for intricate scenarios (e.g., Ref-SAV, LongRVOS).

\noindent\textbf{(2) SEG-aware Logit Calibration.} VideoSEG-O3 adopts the implicit latent embedding paradigm. However, standard GRPO~\cite{deepseekr1} optimizes the likelihood of the \texttt{[SEG]} token but overlooks the quality of its latent embedding required for accurate segmentation (Fig.~\ref{fig:motivation}(b)). To bridge this gap, we introduce \textit{SEG-aware Logit Calibration}, which modulates the \texttt{[SEG]} token's logits using pixel-wise segmentation probabilities. This ensures that the RL process synchronously optimizes both the textual reasoning path and the quality of the segmentation embedding.

\noindent\textbf{(3) Decoupled Thinking Trace.} 
Effective RVOS necessitates the synergistic comprehension of temporal dynamics, spatial details, and linguistic semantics. Unlike prior works that rely on holistic but entangled analysis~\cite{omni-r1, lens}, we design a \textit{Decoupled Thinking Trace} that hierarchically decomposes the reasoning process into three structured stages, supported by resolution-adaptive visual inputs. As shown in Fig.~\ref{fig:overall_motivation} (\textbf{Top}), the MLLM ingests low-resolution frames to capture global temporal context, while utilizing high-resolution keyframes for spatial precision. The reasoning trajectory is strictly organized into: \textit{(i) Temporal Understanding}, \textit{(ii) Spatial Detail Capturing}, and \textit{(iii) Expression Parsing}.

\noindent In summary, our key contributions are as follows:
\noindent \textbf{(1)} We propose \textbf{VideoSEG-O3}, the first multi-turn RL framework for RVOS. By incorporating a \textit{Multi-turn Temporal-Spatial CoT}, it shifts the paradigm from passive one-step inference to iterative visual exploration, enabling the model to autonomously perceive fine-grained details.
\noindent \textbf{(2)} We introduce \textit{SEG-Aware Logit Calibration}, which injects pixel-wise segmentation feedback directly into the \texttt{[SEG]} token's policy.
\noindent \textbf{(3)} We construct \textbf{VTS-CoT}, a cold-start dataset featuring structured reasoning trajectories across temporal, spatial, and linguistic dimensions.
\noindent \textbf{(4)} \textbf{VideoSEG-O3} achieves advanced performance across 8 RVOS benchmarks. Remarkably, even with a compact model scale (4B), it surpasses previous SOTA methods, achieving gains of +6.1\% on the long-video task LongRVOS, +4.2\% on the motion-centric MeViS, and +4.0\% on the reasoning benchmark ReVOS.

\begin{figure*}
    \centering
    \includegraphics[width=1.0\textwidth]{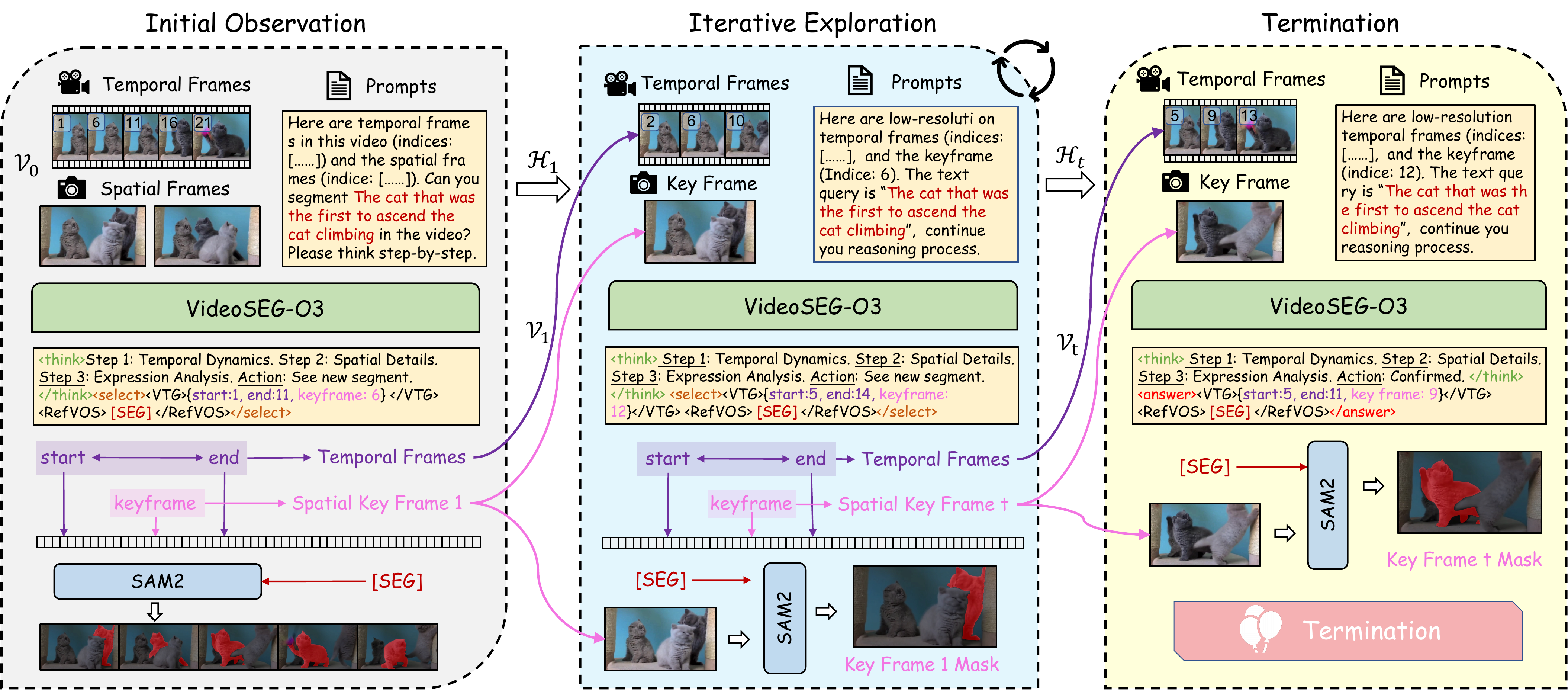} 
    \caption{Overall pipeline of \textbf{VideoSEG-O3}. (1) \textbf{Initial Observation}: The model ingests global temporal frames and uniformly sampled spatial frames as visual inputs. (2) \textbf{Iterative Exploration}: The model iteratively refines its reasoning by taking the temporal interval and keyframe index selected via \texttt{<select>} in the previous turn as input. (3) \textbf{Termination}: Upon generating the \texttt{<answer>} token, the process terminates, and the \texttt{[SEG]} token yields the segmentation mask for the target keyframe.}
    \label{fig:pipeline}
    % \vspace{-5pt}
\end{figure*}

\section{Related Works}

\noindent \textbf{Supervised Fine-Tuning Approaches.}
SFT serves as the dominant paradigm for adapting MLLMs to RVOS~\cite{trackgpt,villa,glus,thinkvideo,videoglamm}.
Pioneering efforts, such as VISA~\cite{visa}, combine MLLMs for temporal reasoning with external object trackers~\cite{xmem} for mask propagation. 
VideoLISA~\cite{videolisa} advances this by introducing a sparse-dense sampling strategy and a \textit{One-Token-Seg-All} paradigm.
More recently, VRS-HQ~\cite{gong2025devil} enhances temporal consistency via token fusion and occlusion-aware keyframe selection using SAM2~\cite{sam2}, while MomentSeg~\cite{momentseg} improves RVOS performance by incorporating temporal sentence grounding capabilities to refine keyframe identification. 
Furthermore, Sa2VA and UniPixel~\cite{sa2va,liu2025unipixel} establish a unified framework for multimodal dense localization and visual-prompt-based captioning tasks, achieving significant performance improvements. 
However, these methods lack an explicit reasoning trace to bridge the linguistic-visual gap and often neglect joint spatiotemporal reasoning. To address this, we propose \textbf{VideoSEG-O3}, which systematically integrates a \textit{Multi-turn Temporal-Spatial Chain-of-Thought} to enhance both reasoning depth and segmentation quality.

\noindent \textbf{Reinforcement Learning Approaches.}
RL has established itself as a potent paradigm for enhancing reasoning capabilities in LLMs, particularly via GRPO~\cite{deepseekr1}, and has recently seen preliminary exploration in reasoning segmentation tasks.
In the \textbf{image domain}, existing methods~\cite{visionreasoner,coprs,sam-r1,pixelthink} leverage specific rewards to improve both image-level reasoning and segmentation quality. Representative approaches like SegZero~\cite{segzero} adopt a box-mediated paradigm, utilizing predicted bounding boxes to prompt the SAM model for segmentation. Conversely, LENS~\cite{lens} introduces learnable tokens to facilitate end-to-end optimization, while R-Sa2VA~\cite{vrt} retains the \texttt{[SEG]} token paradigm~\cite{lisa} to trigger segmentation via a latent embedding.
Transitioning to the \textbf{video domain}, recent works~\cite{omni-r1, li2025revseg} instruct LLMs to generate explicit textual prompts (e.g., bounding boxes or points) via a Chain-of-Thought process, which subsequently guide the SAM2 for mask propagation. 
However, these methods overlook the potential of multi-turn interaction and supplementary visual cues for iterative refinement. To address this, \textbf{VideoSEG-O3} introduces autonomous \textit{Multi-turn Spatiotemporal Exploration} and \textit{SEG-aware Logit Calibration}, enabling direct optimization of the implicit embedding.

\section{VideoSEG-O3}

\begin{figure*}[t]
    \centering
    \includegraphics[width=1.0\textwidth]{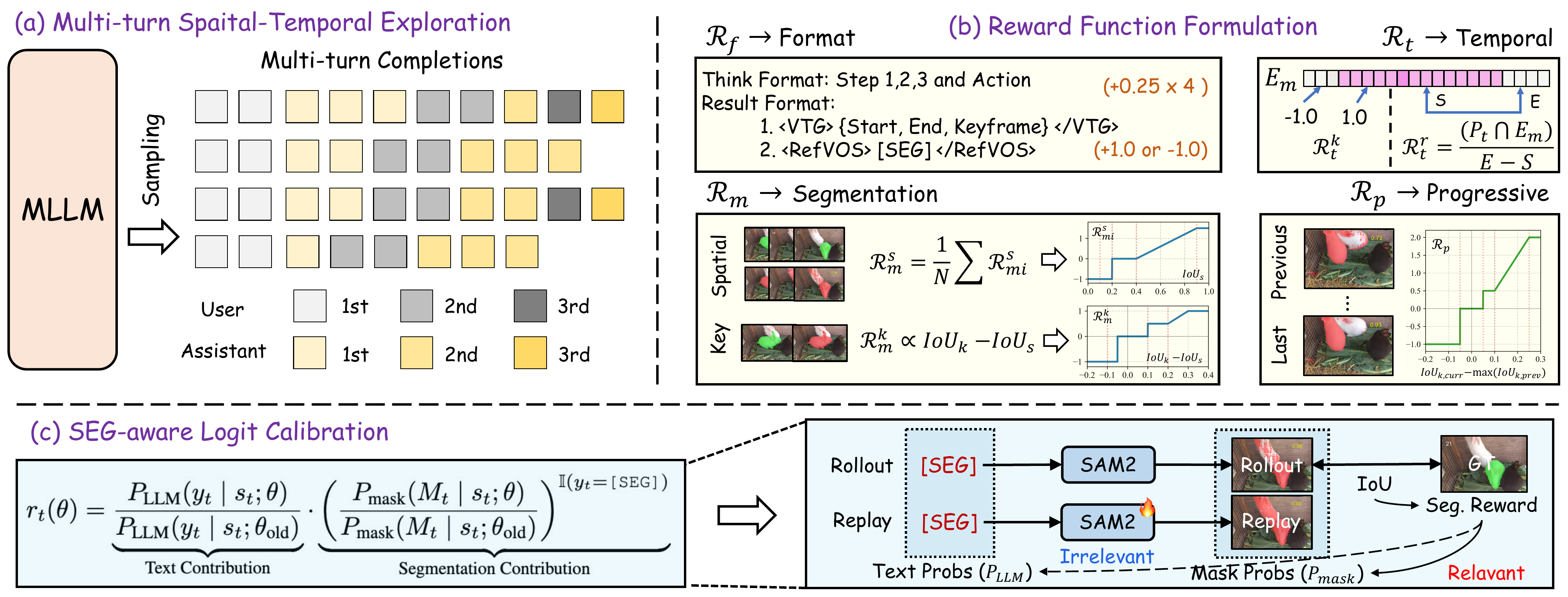} 
  \caption{The overall framework of VideoSEG-O3. 
(a) Multi-turn temporal-spatial exploration (Sec.~\ref{subsec:task_formulation}).
(b) A composite reward design guiding reinforcement learning via format ($\mathcal{R}_f$), temporal ($\mathcal{R}_t$), segmentation ($\mathcal{R}_m$), and progressive ($\mathcal{R}_p$) rewards (Sec.~\ref{subsec:reward_design}).
(c) A calibration strategy that aligns the latent \texttt{[SEG]} token representation with pixel-level mask confidence (Sec.~\ref{subsec:seg_aware_logit_calibration}).
}
\label{fig:framework}
\end{figure*}

\subsection{Task Formulation}
\label{subsec:task_formulation}

In this work, we formulate VideoSEG-O3 as a sequential decision-making process that emulates human visual cognition, iteratively attending to fine-grained video temporal-spatial fragments to achieve precise segmentation. The optimization framework targets two core capabilities: \textit{(1) Active Spatiotemporal Exploration}, enabling the model to identify informative temporal intervals and spatial keyframes; and \textit{(2) Implicit Representation Refinement}, optimizing the \texttt{[SEG]} token embedding for high-fidelity mask generation.
Specifically, the reasoning process begins with a global perception phase, where the model analyzes the full video context alongside sparse spatial cues. This global understanding guides the generation of a \texttt{<select>} token, triggering the active selection of critical temporal intervals and keyframes for detailed inspection. In subsequent iterations, the visual input is dynamically augmented with these refined spatiotemporal details. This cycle continues until the model determines that sufficient evidence has been gathered, signaling termination via the \texttt{<answer>} token to produce the final segmentation mask. As illustrated in Fig.~\ref{fig:pipeline}, this process is formally defined as a Markov Decision Process (MDP) specified by the tuple $(\mathcal{S}, \mathcal{A}, \mathcal{T}, \mathcal{R})$:

\begin{itemize}[leftmargin=*, nosep, itemsep=2pt]
    \item \textbf{State $s \in \mathcal{S}$:} Represents the evolving observational status. Unlike traditional approaches that process a static video representation, our state is dynamic. We define the state at step $t$ as a tuple $s_t = (\mathcal{V}_t, Q, \mathcal{H}_{t-1})$, where $Q$ is the textual query and $\mathcal{H}_{t-1}$ denotes the interaction history context. Crucially, $\mathcal{V}_t$ represents the current visual observation: at $t=0$, $\mathcal{V}_0$ comprises globally sampled frames for initial context; for $t>0$, $\mathcal{V}_t$ is actively updated to include \textit{low-resolution temporal frames} and a \textit{high-resolution keyframe}, derived from the previous action.
    \item \textbf{Action $a \in \mathcal{A}$:} Refers to the decision made by the Policy. The action space consists of two distinct operations: (1) \textit{Exploration}, where the model outputs a \texttt{<select>} token followed by a specific time range and keyframe index to request refined visual details; and (2) \textit{Termination}, where the model outputs an \texttt{<answer>} token followed by the \texttt{[SEG]} token to generate the final segmentation mask.
    \item \textbf{Transition Function $\mathcal{T}: \mathcal{S} \times \mathcal{A} \rightarrow \mathcal{S}$:} Defines the update mechanism for the observation state. If $a_t$ is an exploration action, $\mathcal{T}$ acts as a sampler that retrieves data from the raw video. Specifically, it extracts the requested temporal segment at \textit{low resolution} to maintain motion context, while simultaneously extracting the target keyframe at \textit{high resolution} to capture fine-grained spatial details, forming the new visual input $\mathcal{V}_{t+1}$ for the next iteration.
    \item \textbf{Reward Function $\mathcal{R}: \mathcal{S} \times \mathcal{A} \rightarrow \mathbb{R}$:} Evaluates the quality of the reasoning and segmentation. The reward function incorporates multiple components, including format, temporal, segmentation, and progressive reward (Sec.~\ref{subsec:reward_design}).
\end{itemize}

\subsection{SEG-aware Logit Calibration}
\label{subsec:seg_aware_logit_calibration}

Standard GRPO~\cite{deepseekr1} optimizes the policy $\pi_\theta$ via an importance sampling objective derived exclusively from textual token sequences. For a sequence $y = \{y_1, \dots, y_T\}$, the importance ratio at token $t$ is typically defined solely on the text distribution:
\begin{equation}
\setlength{\abovedisplayskip}{5pt}
\setlength{\belowdisplayskip}{5pt}
    \log r_t(\theta) = \log P_{\text{LLM}}^\theta(y_t) - \log P_{\text{LLM}}^{\theta_{\text{old}}}(y_t).
\end{equation}
While effective for linguistic alignment, this formulation presents a structural limitation in segmentation tasks. Standard GRPO treats the \texttt{[SEG]} token identically to discrete textual tokens, deriving the policy gradient strictly from language head logits. This creates a \textit{gradient disconnect}: the mask decoder is differentiably detached from the policy's estimation. Moreover, as illustrated in Fig.~\ref{fig:motivation}(b), the prediction probability of the \texttt{[SEG]} token is irrelevant to the reward derived from mask quality.
To address this misalignment, we propose \textit{SEG-aware Logit Calibration}, as illustrated in Fig.~\ref{fig:framework}(c). We reformulate the segmentation process as a composite probabilistic event that injects mask confidence into textual probabilities. Specifically, we calibrate the policy's probability for the \texttt{[SEG]} token by directly injecting the pixel-wise confidence from the mask decoder:
\begin{equation}
\setlength{\abovedisplayskip}{5pt}
\setlength{\belowdisplayskip}{5pt}
\begin{split}
    \log \tilde{\pi}_\theta(y_t, M_t | & s_t)   = \, \log P_{\text{LLM}}(y_t | s_t) \\
    & + \mathbb{I}(y_t = \texttt{[SEG]}) \cdot \log P_{\text{mask}}(M_t | \mathbf{h}_t).
\end{split}
\label{eq:calibrated_policy}
\end{equation}
To explicitly quantify the segmentation probability, we treat the binarized mask $M_t$ obtained during the rollout phase as the sampled segmentation outcome. We define $\log P_{\text{mask}}$ as the spatial average of the log-likelihood that the mask decoder generates this specific rollout mask. Crucially, this probability is conditioned on $\mathbf{h}_t$, the hidden state representation of the \texttt{[SEG]} token at step $t$:
\begin{equation}
\setlength{\abovedisplayskip}{5pt}
\setlength{\belowdisplayskip}{5pt}
P_{\text{mask}}(M_t | \mathbf{h}_t) = \frac{1}{HW} \sum_{i,j} \mathcal{P}\left( m_{i,j} \mid \Phi(\mathbf{h}_t)_{i,j} \right).
\label{eq:mask_prob}
\end{equation}
Here, $m_{i,j} \in \{0, 1\}$ represents the pixel value from $M_t$, and $\Phi(\mathbf{h}_t)$ denotes the predicted mask probabilities projected from the hidden states $\mathbf{h}_t$. $\mathcal{P}$ is computed via negative binary cross-entropy, effectively measuring the consistency between the model's replay and rollout processes. 

By substituting Eq.~\ref{eq:calibrated_policy} into the GRPO objective, the importance ratio $r_t(\theta)$ becomes sensitive to visual fidelity. The log-ratio decomposes as:
\begin{equation}
\setlength{\abovedisplayskip}{5pt}
\setlength{\belowdisplayskip}{5pt}
\begin{split}
    & \log r_t(\theta) =  \underbrace{\left( \log P_{\text{LLM}}^\theta(y_t) - \log P_{\text{LLM}}^{\theta_{\text{old}}}(y_t) \right)}_{\text{Text Contribution}} + \\
    & \mathbb{I}(y_t = \texttt{[SEG]})  \underbrace{\left( \log P_{\text{mask}}^\theta(M_t) - \log P_{\text{mask}}^{\theta_{\text{old}}}(M_t) \right)}_{\text{Segmentation Contribution}}.
\end{split}
\label{eq:calibrated_ratio}
\end{equation}
This formulation establishes a pathway from the reward signal to the mask decoder. The gradient update can be modulated by the relative improvement in pixel-wise certainty. This ensures the policy is penalized not only for textual semantic drift but also for spatial uncertainty.

\subsection{Reward and Objectives}

\paragraph{Reward Functions Formulation.}
\label{subsec:reward_design}
To incentivize structured reasoning and high-fidelity grounding, we formulate a composite reward, as illustrated in Fig.~\ref{fig:framework}(b). Detailed descriptions are provided in Appendix~\ref{app:reward_design}. We adopt an episodic reward, and the corresponding ablation studies and detailed analyses are presented in Appendix~\ref{app:reward_ablation}.
\begin{itemize}[leftmargin=*, nosep, itemsep=2pt]
    \item \textbf{Format Reward ($\mathcal{R}_f$):} We enforce strict compliance in two aspects: \textit{Think Format} (ensuring logical CoT steps) and \textit{Result Format} (validating syntactic correctness).

    \item \textbf{Temporal Reward ($\mathcal{R}_t$):} We evaluate localization accuracy through keyframe alignment and temporal precision:
    \begin{equation}
        \setlength{\abovedisplayskip}{5pt}
        \setlength{\belowdisplayskip}{5pt}
        \mathcal{R}_t^k \propto (-1)^{\mathds{1}(k \notin G_t)},
        \quad \mathcal{R}_t^r \propto \frac{|P_t \cap G_t|}{|P_t|},
    \end{equation}
    where $k$, $P_t$, and $G_t$ denote the predicted keyframe, predicted range, and ground truth, respectively.

    \item \textbf{Segmentation Quality Reward ($\mathcal{R}_m$):} We employ reward that correlate with mask quality and keyframe selection superiority. Specifically, $\mathcal{R}_m^s$ represents the cumulative frame-level segmentation quality, and $\mathcal{R}_m^k$ quantifies the relative advantage of selected keyframe over manually sampled spatial frames, formulated as:
    \begin{equation}
        \setlength{\abovedisplayskip}{5pt}
        \setlength{\belowdisplayskip}{5pt}
        \mathcal{R}_m^s \propto \frac{1}{N} \sum_{i=1}^{N} \text{IoU}_i, \quad \mathcal{R}_m^k \propto (\text{IoU}_k - \text{IoU}_s).
    \end{equation}

    \item \textbf{Progressive Reward ($\mathcal{R}_p$):} To incentivize continuous refinement and penalize redundant interactions, we reward the policy only when the current mask quality exceeds the historical maximum. Conversely, a penalty is applied for ineffective turns, formulated as:
    \begin{equation}
        \setlength{\abovedisplayskip}{5pt}
        \setlength{\belowdisplayskip}{5pt}
        \mathcal{R}_p \propto \left( \text{IoU}_{k, t} - \max_{j < t}(\text{IoU}_{k, j}) \right).
    \end{equation}
\end{itemize}

\paragraph{RL Training Objectives.}
\label{subsec:train_objective}
To ensure stable policy optimization and prevent the degradation of segmentation capabilities during exploration, we formulate a hybrid objective function. Relying exclusively on sparse rewards can induce optimization instability, particularly when sampled generations yield flat reward distributions (i.e., vanishing advantage). To mitigate this, we introduce an auxiliary dense supervision signal. Analogous to the KL-divergence constraint in RLHF, this term strictly anchors the latent \texttt{[SEG]} representation to the GT masks. The total loss is defined as:
\begin{equation}
    \setlength{\abovedisplayskip}{5pt}
    \setlength{\belowdisplayskip}{5pt}
    \mathcal{L}(\theta) = \mathcal{L}_{GRPO}(\tilde{\pi}_\theta) + \lambda_{seg} \underbrace{(\mathcal{L}_{spatial} + \mathcal{L}_{key})}_{\text{Joint Segmentation Loss}},
\end{equation}
where $\mathcal{L}_{GRPO}$ maximizes the expected reward using the calibrated policy $\tilde{\pi}_\theta$. The auxiliary segmentation loss comprises $\mathcal{L}_{spatial}$, computed on the spatial frames, and $\mathcal{L}_{key}$, computed on the keyframe. This dual-objective mechanism ensures that the model maintains robust pixel-level decoding capabilities even during the early stages of reasoning exploration. The $\lambda_{seg}$ balances the trade-off between semantic reasoning exploration and spatial grounding stability.

\subsection{Multi-turn Temporal-Spatial Chain-of-Thought}

\paragraph{VTS-CoT Dataset.}
We construct the cold-start CoT dataset through a four-stage pipeline: \textit{(1) Data selection}, \textit{(2) Temporal labeling}, \textit{(3) Candidate generation}, and \textit{(4) Decoupled CoT synthesis}, resulting in more than 6K samples. For the detailed construction process and the pipeline illustration, please refer to Appendix~\ref{app:vtscot}.

\paragraph{Decoupled Thinking Trace.}
To facilitate structured reasoning, we design a \textit{Decoupled Thinking Trace} that hierarchically organizes inference into three stages (Fig.~\ref{fig:overall_motivation}, \textbf{Top}): \textit{(1) Temporal Understanding} using global low-resolution frames, \textit{(2) Spatial Description} via high-resolution spatial frames or keyframe, and \textit{(3) Expression Parsing}. This decoupled design and resolution-adaptive workflow ensure the efficient utilization of input tokens.

\section{Experiments}
\label{sec:experiment}

\begin{table*}[t]
  \centering
  \caption{Main results on referring video object segmentation benchmarks. We compare \textbf{VideoSEG-O3} with specialized RVOS and MLLM-based methods. Best results are \textbf{bolded}, and the second best results are \underline{underlined}.}
  \label{tab:refvos_main_results}
  \resizebox{1.0\linewidth}{!}{
    \begin{tabular}{l|ccc | ccc | ccc | ccc | c}
      \toprule
      \multirow{2}{*}{Method} & \multicolumn{3}{c|}{MeViS} & \multicolumn{3}{c|}{Ref-Youtube-VOS} & \multicolumn{3}{c|}{Ref-DAVIS17} & \multicolumn{3}{c|}{Ref-SAV} & \multicolumn{1}{c}{Long-RVOS} \\
      \cmidrule(lr){2-4} \cmidrule(lr){5-7} \cmidrule(lr){8-10} \cmidrule(lr){11-13} \cmidrule(lr){14-14}
      & $\mathcal{J}\&\mathcal{F}$ & $\mathcal{J}$ & $\mathcal{F}$ & $\mathcal{J}\&\mathcal{F}$ & $\mathcal{J}$ & $\mathcal{F}$ & $\mathcal{J}\&\mathcal{F}$ & $\mathcal{J}$ & $\mathcal{F}$ & $\mathcal{J}\&\mathcal{F}$ & $\mathcal{J}$ & $\mathcal{F}$ & $\mathcal{J}\&\mathcal{F}$ \\ 
      \midrule
      \rowcolor{gray!10} \multicolumn{14}{c}{\textbf{Specialized Methods}} \\
      SAMWISE~\citep{samwise}         & 49.5 & 46.6 & 52.4 & 69.2 & 67.8 & 70.6 & 70.6 & 67.4 & 74.5 & -- & -- & -- & 35.6 \\
      ReferDINO~\citep{referdino}     & 49.3 & 44.7 & 53.9 & 69.3 & 67.0 & 71.5 & 68.9 & 65.1 & 72.9 & -- & -- & -- & 48.7 \\
      ReferMo~\citep{longrvos}        & -- & -- & -- & -- & -- & -- & -- & -- & -- & -- & -- & -- & 51.3 \\
      \midrule
      \rowcolor{gray!10} \multicolumn{14}{c}{\textbf{MLLM-based Methods}} \\
      LISA-7B~\citep{lisa}               & 37.2 & 35.1 & 39.4 & 53.9 & 53.4 & 54.3 & 64.8 & 62.2 & 67.3 & -- & -- & -- & -- \\
      VideoLISA-3.8B~\citep{videolisa}   & 44.4 & 41.3 & 47.6 & 63.7 & 61.7 & 65.7 & 68.8 & 64.9 & 72.7 & -- & -- & -- & 33.1 \\
      VISA-7B~\citep{visa}               & 43.5 & 40.7 & 46.3 & 61.5 & 59.8 & 63.2 & 69.4 & 66.3 & 72.5 & 11.8 & 13.2 & 11.3 & -- \\
      GLUS-7B~\citep{glus}               & 51.3 & 48.5 & 54.2 & 67.3 & 65.5 & 69.0 & --    & --    & --    & -- & -- & -- & 36.6 \\
      InstructSeg-3B~\citep{instructseg} & -- & -- & -- & 67.5 & 65.4 & 69.5 & 71.1 & 67.3 & 74.9 & -- & -- & -- & -- \\
      Sa2VA-8B~\citep{sa2va}             & 46.2 & --    & --    & 70.0 & --    & --    & 70.0 & --    & --    & 50.0 & 48.3 & 51.7 & --  \\
      UniPixel-7B~\citep{liu2025unipixel}& \underline{55.8} & \underline{53.2} & 58.3 & \underline{71.0} & \underline{69.5} & 72.4 & 76.4 & 72.7 & 80.1 & -- & -- & -- & -- \\
    %   ReVSeg-7B~\citep{li2025revseg}  & \underline{59.8} & \underline{56.1} & \underline{63.4}  & \underline{73.1} & \underline{71.1} & \underline{75.2} &  \underline{80.8} &  \underline{77.4} & \underline{84.1} & -- & -- & -- & -- \\
      \midrule
      \rowcolor{citeblue!5} \textbf{VideoSEG-O3-2B (Ours)} & 55.6 & 52.5 & \underline{58.6} & 70.5 & 68.3 & \underline{72.7} & \textbf{80.0} & \textbf{76.0} & \textbf{84.0} & \underline{62.9} & \underline{61.8} & \underline{64.0} & \underline{54.8} \\
      \rowcolor{citeblue!5} \textbf{VideoSEG-O3-4B (Ours)} & \textbf{60.0} & \textbf{57.0} & \textbf{63.0} & \textbf{74.1} & \textbf{71.5} & \textbf{76.6} & \underline{79.4} & \underline{75.5} & \underline{83.2} & \textbf{65.5} & \textbf{64.3} & \textbf{66.7} & \textbf{57.4} \\
      \bottomrule
    \end{tabular} 
  }
\end{table*}

\begin{table*}[t]
  \centering
  \caption{Main results on reasoning video object segmentation. We compare \textbf{VideoSEG-O3} with state-of-the-art SFT-based and RL-based methods. Our model establishes advanced performance across both in-domain (ReVOS) and out-domain (ReasonVOS, GroundMoRe) zero-shot benchmarks. Best results are \textbf{bolded}, and the second best results are \underline{underlined}.}
  \label{tab:reasonvos_main_results}
  \resizebox{1.0\textwidth}{!}{
    \begin{tabular}{l|ccc ccc ccc | c c}
      \toprule
      \multirow{3}{*}{Method} & \multicolumn{9}{c|}{ReVOS (In-Domain)} & \multicolumn{2}{c}{Out-Domain (ZS)} \\ 
      \cmidrule(lr){2-10} \cmidrule(lr){11-12}
      & \multicolumn{3}{c}{Referring} & \multicolumn{3}{c}{Reasoning} & \multicolumn{3}{c|}{Overall} & \multirow{2}{*}{ReasonVOS} & \multirow{2}{*}{GroundMoRe} \\ 
      % \cmidrule(lr){2-4} \cmidrule(lr){5-7} \cmidrule(lr){8-10}
      & $\mathcal{J}\&\mathcal{F}$ & $\mathcal{J}$ & $\mathcal{F}$ & $\mathcal{J}\&\mathcal{F}$ & $\mathcal{J}$ & $\mathcal{F}$ & $\mathcal{J}\&\mathcal{F}$ & $\mathcal{J}$ & $\mathcal{F}$ & & \\
      \midrule
      \rowcolor{gray!10} \multicolumn{12}{c}{\textbf{SFT-based Methods}} \\
      TrackGPT-13B~\citep{trackgpt} & 49.5 & 48.3 & 50.6 & 40.5 & 38.1 & 42.9 & 45.0 & 43.2 & 46.8 & -- & -- \\
      VISA-13B~\citep{visa}         & 57.4 & 55.6 & 59.1 & 44.3 & 42.0 & 46.7 & 50.9 & 48.8 & 52.9 & -- & 5.31 \\
      VideoLISA-3.8B~\citep{videolisa} & -- & -- & -- & -- & -- & -- & -- & -- & -- & 47.5 & -- \\
      HyperSeg-3B~\citep{hyperseg}  & 58.5 & 56.0 & 60.9 & 53.0 & 50.2 & 55.8 & 55.7 & 53.1 & 58.4 & -- & -- \\
      VRS-HQ-7B~\cite{gong2025devil}& 62.1 & 59.8 & 64.5 & 56.1 & 53.5 & 58.7 & 59.1 & 56.6 & 61.6 & -- & -- \\
      MoRA-13B~\cite{groundmore}   & -- & -- & -- & -- & -- & -- & -- & -- & -- & -- & 23.1 \\
      UniPixel-7B~\citep{liu2025unipixel}& 65.8 & 63.9 & 67.8 & 61.5 & \underline{59.4} & 63.7 & 63.7 & 61.7 & 65.7 & -- & -- \\
      \midrule
      \rowcolor{gray!10} \multicolumn{12}{c}{\textbf{RL-based Methods}} \\
      % Omni-R1-11B~\citep{omni-r1}   & 64.1 & -- & -- & 53.7 & -- & -- & 58.9 & -- & -- & -- & -- \\
      Veason-R1-7B~\citep{Veason_R1}& 63.6 & 60.7 & 66.5 & 59.0 & 55.8 & 62.2 & 61.3 & 58.2 & 64.4 & 59.9 & -- \\
    %   OneThinker-8B~\citep{onethinker}   & -- & -- & -- & -- & -- & -- & -- & -- & -- & 54.9 & -- \\
      VideoSeg-R1-7B~\citep{videoseg-r1}& -- & -- & -- & -- & -- & -- & 61.1 & 58.2 & 64.0 & -- & -- \\
    %   OneThinker-8B~\citep{onethinker}   & -- & -- & -- & -- & -- & -- & -- & -- & -- & 54.9 & -- \\
      \midrule
      \rowcolor{citeblue!5} \textbf{VideoSEG-O3-2B (Ours)} & \underline{67.5} & \underline{65.0} & \underline{70.0} & \underline{62.0} & 59.0 & \underline{65.0} & \underline{64.8} & \underline{62.0} & \underline{67.5} & \underline{60.2} & \underline{29.1} \\
      \rowcolor{citeblue!5} \textbf{VideoSEG-O3-4B (Ours)} & \textbf{70.3} & \textbf{67.7} & \textbf{72.9} & \textbf{65.1} & \textbf{62.4} & \textbf{67.9} & \textbf{67.7} & \textbf{65.0} & \textbf{70.4} & \textbf{62.9} & \textbf{31.9} \\
      \bottomrule
    \end{tabular} 
  }
\end{table*}

\begin{table*}[t]
  \centering
%    \vspace{-2mm}
  \caption{Comparison of model results across image segmentation tasks. We compare \textbf{VideoSEG-O3} with both large-scale MLLM-based methods and lightweight specialist methods. The metric is cIoU. Best results are \textbf{bolded}, and the second best results are \underline{underlined}.}
    % \vspace{-2mm}
    \resizebox{0.9\textwidth}{!}{ 
  \begin{tabular}{l|ccc|ccc|cc|cc}
    \toprule
    \multirow{2}{*}{Method}  & \multicolumn{3}{c|}{RefCOCO} & \multicolumn{3}{c|}{RefCOCO+} & \multicolumn{2}{c|}{RefCOCOg} & \multicolumn{2}{c}{ReasonSeg (ZS)} \\
        \cmidrule(lr){2-4} \cmidrule(lr){5-7} \cmidrule(lr){8-9} \cmidrule(lr){10-11}
        & val & testA & testB & val & testA & testB & val(U) & test(U) & val & test \\
        \midrule
        \rowcolor{gray!10} \multicolumn{11}{c}{\textbf{Specialist Methods}} \\
        ReLA~\citep{gres} & 73.8 & 76.5 & 70.2 & 66.0 & 71.0 & 57.7 & 65.0 & 66.0 & -- & -- \\
        EEVG~\citep{eevg} & 79.5 & 80.9 & 77.4 & 71.9 & 76.7 & 66.3 & 73.6 & 73.5 & -- & -- \\
        PropVG~\citep{propvg} & \textbf{82.0} & \textbf{83.6} & \underline{80.0} & \textbf{77.1} & \underline{79.8} & \textbf{72.2} & \underline{77.0} & \underline{77.7} & -- & -- \\
        InstanceVG~\citep{instancevg} & \underline{81.4} & 83.1 & 79.3 & \underline{76.6} & 79.5 & \underline{71.6} & 75.9 & 76.6 & -- & -- \\
        \midrule
        \rowcolor{gray!10} \multicolumn{11}{c}{\textbf{MLLM-based Methods}} \\
        % PixelLM-7B~\citep{pixellm} & 73.0 & 76.5 & 68.2 & 66.3 & 71.7 & 58.3 & 69.3 & 70.5 & -- & -- \\
        LISA-7B~\citep{lisa} & 74.9 & 79.1 & 72.3 & 65.1 & 70.8 & 58.1 & 67.9 & 70.6 & 61.3 & \textbf{62.9} \\
        GLaMM-7B~\citep{glamm} & 79.5 & \underline{83.2} & 76.9 & 72.6 & 78.7 & 64.6 & 74.2 & 74.9 & -- & -- \\
        VisionLLMv2-7B~\citep{visionllmv2} & 79.2 & 82.3 & 77.0 & 68.9 & 75.8 & 61.8 & 73.3 & 74.8 & 56.9 & 48.3 \\
        OMG-LLaVA-7B~\citep{omg-llava} & 75.6 & 77.7 & 71.2 & 65.6 & 69.7 & 58.9 & 70.7 & 70.2 & -- & -- \\
        % LaSagnA-7B~\citep{lasagna} & 76.8 & 78.7 & 73.8 & 66.4 & 70.6 & 60.1 & 70.6 & 71.9 & 48.8 & 47.2 \\
        VISA-7B~\citep{visa} & 72.4 & 75.5 & 68.1 & 59.8 & 64.8 & 53.1 & 65.5 & 66.4 & 52.7 & 57.8 \\
        VRS-HQ-7B~\citep{gong2025devil} & 73.5 & 77.5 & 69.5 & 61.7 & 67.6 & 54.3 & 66.7 & 67.5 & 51.8 & 52.9 \\
        Sa2VA-4B~\citep{sa2va} & 78.9 & -- & -- & 71.7 & -- & -- & 74.1 & -- & -- & -- \\
        MomentSeg-7B~\citep{momentseg} & 77.8 & 80.4 & 73.9 & 68.2 & 74.5 & 57.8 & 70.8 & 72.3 & -- & -- \\
        \midrule
        \rowcolor{gray!10} \multicolumn{11}{c}{\textbf{RL-based Methods}} \\
        Seg-Zero-7B~\citep{segzero} & -- & 80.3 & -- & -- & 76.2 & -- & -- & 72.6 & 52.6 & 52.4 \\
        PixelThink-7B~\citep{pixelthink} & -- & 79.3 & -- & -- & 74.8 & -- & -- & 73.9 & 62.7 & 55.8 \\
        VideoSeg-R1-7B~\citep{videoseg-r1} & 78.2 & 82.3 & 75.1 & 71.8 & 76.1 & 64.7 & 73.1 & 74.1 & \textbf{68.2} & -- \\
        \midrule
        \rowcolor{citeblue!5} \textbf{VideoSEG-O3-4B (Ours)} & 80.9 & 82.5 & \textbf{80.5} & 75.5 & \textbf{79.9} & 70.6 & \textbf{78.9} & \textbf{79.0} & \underline{67.7} & \underline{61.5} \\
    \bottomrule
  \end{tabular}
    }
  \label{tab:image_segmentation}
%    \vspace{-4mm}
\end{table*}

\subsection{Implementation Details}
\label{subsec:implementation_details}

\paragraph{Architecture and Setup.}
We construct our baseline by integrating Qwen3-VL~\citep{Qwen3-VL} with the SAM2~\citep{sam2}. Following prior approaches~\citep{sa2va}, segmentation masks are generated by decoding the hidden state of the special \texttt{[SEG]} token via the SAM2 decoder. As outlined in Table~\ref{tab:training_stages}, the training pipeline of VideoSEG-O3 comprises three distinct stages. During the SFT and CoT stages, we employ LoRA~\citep{lora} to fine-tune the MLLM with a maximum sequence length of 8,192 tokens. Conversely, the RL stage involves full-parameter fine-tuning of the MLLM. All experiments are conducted on 8 NVIDIA H20 GPUs.  The RL stage of VideoSEG-O3 involves 1,000 training steps. This process requires approximately 24 hours for VideoSEG-O3-4B and roughly 16 hours for the VideoSEG-O3-2B variant.

\paragraph{Training Stages.}
We adopt a hierarchical three-stage pipeline to progressively build VideoSEG-O3's capabilities (detailed dataset composition is in Table~\ref{tab:training_stages}). \textit{Stage I (Supervised Fine-Tuning):} This phase establishes core competencies by aligning the \texttt{[SEG]} token with the mask decoder and, crucially, enforcing frame-level \textit{temporal index} understanding to support discrete temporal reasoning. \textit{Stage II (CoT Cold-Start):} Utilizing our constructed VTS-CoT dataset, this stage elicits structured multi-step reasoning and tool usage, standardizing the format for multi-turn interactions. \textit{Stage III (Reinforcement Learning):} Finally, we optimize the policy for iterative exploration, leveraging composite rewards to guide the model in seeking discriminative visual cues and refining segmentation accuracy across turns.

\begin{table*}[t]
\centering
\caption{Ablation study on training stages. All metrics are reported using the $\mathcal{J}\&\mathcal{F}$ score. Incremental gains represent absolute improvements over the previous stage, except for \textit{Avg. Round}, which denotes the percentage reduction. Evaluation is performed using keyframe initialization without post-processing.} 
\label{tab:ablation_stages}
\resizebox{1.0\linewidth}{!}{
  \begin{tabular}{l|cccccc|cc}
  \toprule
  Model & MeViS ($val_u$) & MeViS ($val$) & Ref-Youtube-VOS & ReVOS & Long-RVOS & Ref-SAV & Avg. $\mathcal{J}\&\mathcal{F}$ & Avg. Rounds\\
  \midrule
  VideoSEG-SFT & 56.71 & 47.04 & 67.80 & 59.10 & 47.12 & 51.97 & 54.96 & / \\
  % CoT Stage: Mostly positive, but Ref-SAV drops (-1.86), so it is marked RED
  VideoSEG-CoT & 59.43 {\color{green!60!black}\scriptsize(+2.72)} & 50.55 {\color{green!60!black}\scriptsize(+3.51)} & 70.12 {\color{green!60!black}\scriptsize(+2.32)} & 60.37 {\color{green!60!black}\scriptsize(+1.27)} & 51.82 {\color{green!60!black}\scriptsize(+4.70)} & 50.11 {\color{gray}\scriptsize(-1.86)} & 57.07 {\color{green!60!black}\scriptsize(+2.11)} & 2.79 \\
  % RL Stage: All metrics improve (+), Rounds decrease (efficiency gain), so all GREEN
  VideoSEG-RL  & \textbf{62.43} {\color{green!60!black}\scriptsize(+3.00)} & \textbf{53.22} {\color{green!60!black}\scriptsize(+2.67)} & \textbf{72.52} {\color{green!60!black}\scriptsize(+2.40)} & \textbf{64.24} {\color{green!60!black}\scriptsize(+3.87)} & \textbf{54.66} {\color{green!60!black}\scriptsize(+2.84)} & \textbf{54.64} {\color{green!60!black}\scriptsize(+4.53)} & \textbf{60.29} {\color{green!60!black}\scriptsize(+3.22)} & \textbf{2.23} {\color{green!60!black}\scriptsize($\downarrow$20.1\%)} \\
  \bottomrule
  \end{tabular}
}
\end{table*}

\begin{figure*}[t]
    \centering
    \includegraphics[width=1.0\textwidth]{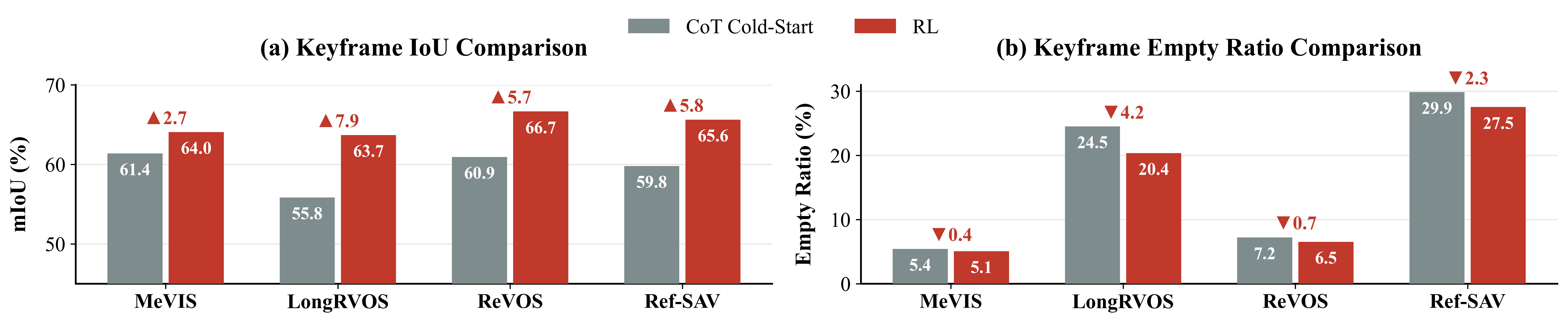} 
    \caption{{Quantitative comparison of RL on keyframe selection.} We compare \textit{keyframe mIoU} (mask quality for initialization) and \textit{keyframe empty ratio} (frequency of selecting frames without the target) across four representative benchmarks.}
    \label{fig:keyframe_shot}
    \vspace{-5pt}
\end{figure*}

\begin{table}[t]
\centering
\caption{Ablation study on RL optimization components. All metrics are reported using the $\mathcal{J}\&\mathcal{F}$ score. All evaluations are conducted without post-processing to maintain consistency.} 
\label{tab:ablation_rl_v1}
\resizebox{0.48\textwidth}{!}{
  \begin{tabular}{l|cc} 
  \toprule
  Model & MeViS ($val_u$) & Ref-SAV \\
  \midrule
  Stage II (Cold-Start) & 59.43 & 50.55 \\
  \midrule
  Baseline (GRPO-RL) & 57.92 {\color{gray}\scriptsize(-1.51)} & 49.78 {\color{gray}\scriptsize(-0.77)} \\
  \midrule
  + SEG-aware Logit Calibration & 60.51 {\color{green!60!black}\scriptsize(+2.59)} & 51.27 {\color{green!60!black}\scriptsize(+1.49)} \\
  \midrule
  \multicolumn{3}{l}{\textit{Auxiliary Segmentation Loss}} \\ 
  \quad + Spatial-wise & 60.71 {\color{green!60!black}\scriptsize(+0.20)} & 51.76 {\color{green!60!black}\scriptsize(+0.49)} \\
  \quad + Keyframe-wise & 61.31 {\color{green!60!black}\scriptsize(+0.80)} & 52.61 {\color{green!60!black}\scriptsize(+1.34)} \\
  + Joint Segmentation Loss & \textbf{61.95} {\color{green!60!black}\scriptsize(+1.44)} & \textbf{53.22} {\color{green!60!black}\scriptsize(+1.95)} \\
  \bottomrule
  \end{tabular}
}
\end{table}

\subsection{Main Results}

\paragraph{Referring Video Object Segmentation.} 
Table~\ref{tab:refvos_main_results} demonstrates that VideoSEG-O3-4B achieves state-of-the-art performance across five RefVOS benchmarks by a substantial margin. It exhibits superior generalization over specialized methods~\cite{samwise,referdino} and dominates MLLM-based methods~\cite{sa2va} on complex tasks, achieving gains of \textbf{+15.5\%} on Ref-SAV and \textbf{+6.1\%} on Long-RVOS. On traditional benchmarks such as MeViS, Ref-YouTube-VOS, and Ref-DAVIS17, VideoSEG-O3 at a 2B scale achieves performance comparable to the previous SOTA UniPixel-7B~\cite{liu2025unipixel}, demonstrating significantly higher parameter efficiency.

\paragraph{Reasoning Video Object Segmentation.}
As shown in Table~\ref{tab:reasonvos_main_results}, VideoSEG-O3 achieves advanced performance in both in-domain and out-domain scenarios. On ReVOS, it surpasses the strongest RL competitor, Veason-R1~\cite{Veason_R1}, by \textbf{+6.4\%} overall, with a notable \textbf{+6.1\%} gain in the \textit{Reasoning} sub-category, validating the efficacy of our VideoSEG-O3 in handling complex linguistic logic. Furthermore, VideoSEG-O3-4B demonstrates robust zero-shot generalization, outperforming the MoRA-13B~\cite{groundmore} on GroundMoRe benchmark by \textbf{+8.8\%}.

\paragraph{Referring/Reasoning Image Segmentation.}
Although VideoSEG-O3 is primarily designed for video scenarios, image-level segmentation data was incorporated during the Supervised Fine-Tuning (SFT) phase. To evaluate whether image-level capabilities are preserved following video-centric training, we conducted benchmarks on standard image datasets. As demonstrated in Table~\ref{tab:image_segmentation}, VideoSEG-O3 maintains competitive performance on image tasks, suggesting that the model effectively generalizes across both static and temporal domains.

\subsection{Ablation Study}

\paragraph{Impact of Training Stages.} Table~\ref{tab:ablation_stages} delineates the performance trajectory across our three-stage pipeline. \textbf{Stage I (SFT)} establishes the foundational grounding proficiency, achieving an average $\mathcal{J}\&\mathcal{F}$ of 54.96\% by enabling the MLLM to generate \texttt{[SEG]} embeddings for the SAM2 decoder. Building upon this, \textbf{Stage II (CoT)} leverages structured reasoning and decoupled visual sampling to enhance spatiotemporal localization. This results in significant absolute gains on complex benchmarks, such as MeViS ($val_u$) (+2.72\%) and Long-RVOS (+4.70\%), raising the overall average to 57.07\%. Although a performance fluctuation is observed on Ref-SAV ($-1.86\%$) due to the distributional shift in VTS-CoT data, \textbf{Stage III (RL)} successfully recovers this drop with a substantial +6.53\% improvement. Notably, the RL stage not only achieves an improved average $\mathcal{J}\&\mathcal{F}$ of 60.29\% but also optimizes inference efficiency, reducing the \textit{Avg. Round} by 20.1\% (2.79 $\rightarrow$ 2.23). This demonstrates that the VideoSEG-O3 achieves superior performance with fewer interaction rounds, indicating a learned ability to precisely sample keyframes essential for target identification.

\paragraph{Refinement of Keyframe Selection via RL.} Fig.~\ref{fig:keyframe_shot} illustrates the comparative impact of reinforcement learning on the model's keyframe selection capability across four benchmarks. We evaluate this refinement through two primary metrics: \textit{keyframe mIoU}, which assesses the model's ability to identify salient frames to provide high-quality initializations for SAM 2 propagation, and \textit{keyframe empty ratio}, which measures the frequency of selecting frames where the target object is absent. Our results demonstrate that the RL-refined model achieves consistently higher mIoU scores across all datasets, signifying a superior capacity for localizing discriminative visual anchors. Furthermore, a substantial reduction in the keyframe empty ratio post-RL indicates that the model autonomously learns to avoid non-informative frames, favoring target-salient views that effectively bridge the gap between linguistic expressions and dense pixel-level grounding.

\begin{figure*}[t]
    \centering
    \includegraphics[width=1.0\textwidth]{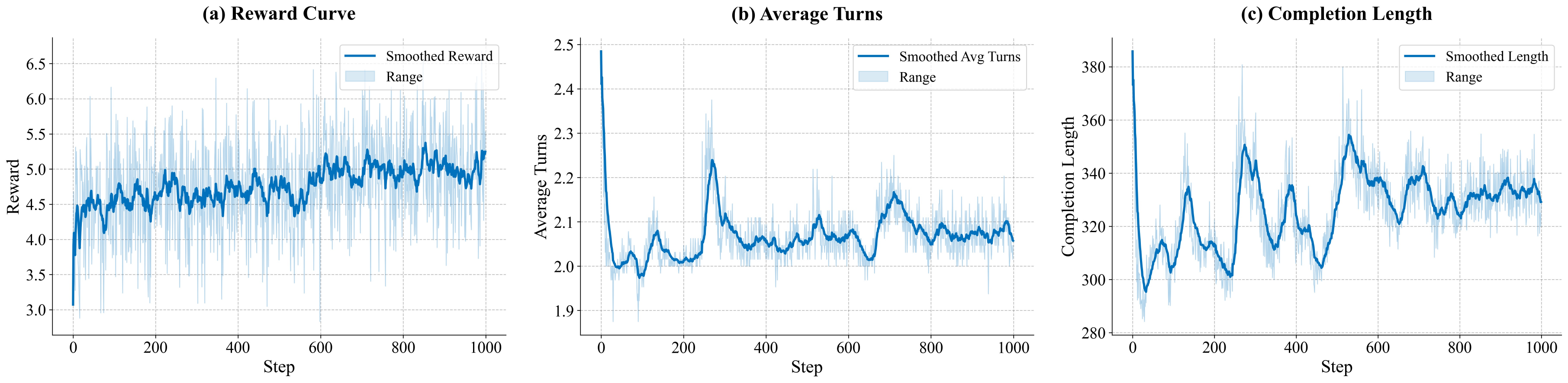} 
    \caption{\textbf{Reinforcement learning training curves.} (a) The smoothed reward curve demonstrates consistent convergence and performance improvement. (b) The average turns indicate the model's transition from fixed reasoning patterns to adaptive multi-turn exploration. (c) The completion length reflects the refinement and stabilization of the generated reasoning trajectories.}
    \label{fig:rl_training_curve}
\end{figure*}

\paragraph{Effectiveness of RL Optimization Components.} 
We investigate the contribution of each RL component in Table~\ref{tab:ablation_rl_v1}. Our findings highlight three key insights: 
(1) Applying standard GRPO solely for text-level optimization leads to performance degradation (59.43\% $\rightarrow$ 57.92\% on MeViS) compared to the Stage II baseline. Fundamentally, the gradients for the SAM2 decoder are obstructed because standard GRPO optimizes the probability of textual token generation. Consequently, an intrinsic disconnect arises between the segmentation-based reward and the text prediction, preventing the reward signal from being effectively attributed to the rollout samples.
(2) The introduction of \textit{SEG-aware Logit Calibration} successfully reverses this trend by aligning the optimization objective with pixel-level fidelity, yielding immediate gains (49.78\%$\rightarrow$51.27\% on Ref-SAV). Furthermore, we observe that this mechanism enhances training stability and mitigates optimization collapse. We hypothesize that this is achieved by aligning the reward signals with the policy logits. Without this calibration, the misalignment between segmentation-related rewards and text probabilities renders the estimation of relative advantage unreliable.
(3) The \textit{Auxiliary Segmentation Loss} further enhances the discriminative capacity of the \texttt{[SEG]} embedding by providing direct supervision to the SAM 2 decoder. While spatial-wise and keyframe-wise losses independently improve performance, their joint application achieves optimal results (60.51\%$\rightarrow$61.95\% on MeViS), demonstrating their complementary roles in stabilizing multi-turn RL training.

\paragraph{Analysis of RL Training Dynamics.}
Fig.~\ref{fig:rl_training_curve} illustrates the optimization progression during the reinforcement learning stage across three key dimensions. First, the \textbf{reward curve} in Fig.~\ref{fig:rl_training_curve}(a) exhibits a steady upward trajectory, confirming the effective convergence of the policy toward higher segmentation fidelity. Second, Fig.~\ref{fig:rl_training_curve}(b) shows that the \textbf{average interaction turns} experience a sharp initial decline. This indicates that the RL process successfully rectifies the rigid patterns inherited from the CoT cold-start phase by eliminating redundant or irrelevant turns. Subsequently, the model enters an exploratory phase where turns slightly increase and fluctuate, eventually stabilizing as the model learns to autonomously allocate necessary reasoning rounds to maximize rewards. Finally, as shown in Fig.~\ref{fig:rl_training_curve}(c), the \textbf{completion length} follows a similar trend to the interaction turns, further validating that VideoSEG-O3 learns to generate more concise and semantically efficient reasoning trajectories through iterative refinement.

\section{Conclusion}
In this work, we presented \textbf{VideoSEG-O3}, a novel multi-turn reinforcement learning framework that fundamentally redefines the paradigm of Reasoning Video Object Segmentation (RVOS) by transitioning from passive, single-step inference to \textit{active, iterative exploration}. 
By mimicking human cognitive strategies, our \textit{Multi-turn Temporal-Spatial Chain-of-Thought} empowers the model to autonomously navigate complex video content through iterative spatio-temporal refocusing. This hierarchical and recursive mechanism ensures that the model maintains a holistic understanding of the video while precisely capturing the target object across dynamic scenes.
Furthermore, the introduction of \textit{SEG-aware Logit Calibration} effectively bridges the optimization gap between discrete textual reasoning and dense pixel-level prediction, enabling synchronized and efficient end-to-end optimization within the RL stage.

\clearpage

\section*{Impact Statement}
This work aims to advance video object segmentation by improving models' ability to reason over temporal, spatial, and linguistic cues through iterative interaction. It may benefit interactive annotation, assistive perception, robotics, and scientific video analysis. However, more capable segmentation systems may also raise privacy, misuse, and bias concerns, especially in surveillance or large-scale video analytics. We encourage deployment-specific evaluation, transparent reporting of limitations and failure cases, and appropriate safeguards such as consent, privacy protection, and human oversight in high-stakes settings.

% Authors are \textbf{required} to include a statement of the potential broader
% impact of their work, including its ethical aspects and future societal
% consequences. This statement should be in an unnumbered section at the end of
% the paper (co-located with Acknowledgements -- the two may appear in either
% order, but both must be before References), and does not count toward the paper
% page limit. In many cases, where the ethical impacts and expected societal
% implications are those that are well established when advancing the field of
% Machine Learning, substantial discussion is not required, and a simple
% statement such as the following will suffice:

% ``This paper presents work whose goal is to advance the field of Machine
% Learning. There are many potential societal consequences of our work, none
% which we feel must be specifically highlighted here.''

% The above statement can be used verbatim in such cases, but we encourage
% authors to think about whether there is content which does warrant further
% discussion, as this statement will be apparent if the paper is later flagged
% for ethics review.

% In the unusual situation where you want a paper to appear in the
% references without citing it in the main text, use \nocite
% \nocite{langley00}

\bibliography{main}
\bibliographystyle{icml2026}

% % %%%%%%%%%%%%%%%%%%%%%%%%%%%%%%%%%%%%%%%%%%%%%%%%%%%%%%%%%%%%%%%%%%%%%%%%%%%%%%%
% % %%%%%%%%%%%%%%%%%%%%%%%%%%%%%%%%%%%%%%%%%%%%%%%%%%%%%%%%%%%%%%%%%%%%%%%%%%%%%%%
% % % APPENDIX
% % %%%%%%%%%%%%%%%%%%%%%%%%%%%%%%%%%%%%%%%%%%%%%%%%%%%%%%%%%%%%%%%%%%%%%%%%%%%%%%%
% % %%%%%%%%%%%%%%%%%%%%%%%%%%%%%%%%%%%%%%%%%%%%%%%%%%%%%%%%%%%%%%%%%%%%%%%%%%%%%%%
\newpage
\appendix
\onecolumn
\section*{Appendix Overview}
This appendix provides supplementary details supporting the main manuscript. The content is organized as follows:
\begin{table}[h]
\centering % 这一行命令让表格在页面中间居中
\renewcommand{\arraystretch}{1.3} % 增加行高，让表格看起来不拥挤
% 使用 tabularx 让表格宽度自适应，p{...} 指定列宽
\begin{tabular}{p{0.05\textwidth} p{0.35\textwidth} p{0.5\textwidth}}
\toprule
\multicolumn{3}{c}{\large \textbf{Appendix Overview}} \\ % 标题跨越三列并居中
\midrule
% 第一行：Implementation Details
& \textbf{\ref{app:implementation}. Implementation Details} & 
\begin{minipage}[t]{\linewidth} % 使用 minipage 包裹列表
    \begin{itemize}[label=$\cdot$, leftmargin=1em, nosep, after=\vspace{0.5em}]
        \item \ref{app:datasets} Datasets and Evaluation Metrics
        \item \ref{app:training} Training Details (SFT, CoT, RL)
        \item \ref{app:inference} Inference Strategy
    \end{itemize}
\end{minipage} \\
\cmidrule{2-3} % 添加分隔线

% 第二行：VTS-CoT Construction
& \textbf{\ref{app:vtscot}. VTS-CoT Construction Details} & 
\begin{minipage}[t]{\linewidth}
    \begin{itemize}[label=$\cdot$, leftmargin=1em, nosep, after=\vspace{0.5em}]
        \item \ref{app:vtscot_pipeline} Pipeline Overview
        \item \ref{app:vtscot_prompts} Prompt Designs
    \end{itemize}
\end{minipage} \\
\cmidrule{2-3}

% 第三行：Additional Methods
& \textbf{\ref{app:methods}. Additional Methods} & 
\begin{minipage}[t]{\linewidth}
    \begin{itemize}[label=$\cdot$, leftmargin=1em, nosep, after=\vspace{0.5em}]
        \item \ref{app:reward_design} VideoSEG-O3 Reward Design Details
    \end{itemize}
\end{minipage} \\
\cmidrule{2-3}

% 第四行：Quantitative Results
& \textbf{\ref{app:results}. Additional Quantitative Results} & 
\begin{minipage}[t]{\linewidth}
    \begin{itemize}[label=$\cdot$, leftmargin=1em, nosep, after=\vspace{0.5em}]
        \item \ref{app:groundmore} Results on GroundMoRe Tasks
    \end{itemize}
\end{minipage} \\
\cmidrule{2-3}

% 第五行：Ablation Studies
& \textbf{\ref{app:ablation}. Additional Ablation Studies} & 
\begin{minipage}[t]{\linewidth}
    \begin{itemize}[label=$\cdot$, leftmargin=1em, nosep, after=\vspace{0.5em}]
        \item \ref{app:post_process} Effect of Post-Processing Strategies
        \item \ref{app:seg_loss} Effectiveness of Joint Segmentation Loss
        \item \ref{app:reward_ablation} Effectiveness of Reward Design
    \end{itemize}
\end{minipage} \\
\cmidrule{2-3}

% 第六行：Qualitative Results
& \textbf{\ref{app:qualitative}. Additional Qualitative Results} & 
\begin{minipage}[t]{\linewidth}
    \begin{itemize}[label=$\cdot$, leftmargin=1em, nosep, after=\vspace{0.5em}]
        \item \ref{app:vis_multiturn} Multi-turn Reasoning Visualizations
    \end{itemize}
\end{minipage} \\
\bottomrule
\end{tabular}
\end{table}

% =========================================================================
% Section A: Implementation Details
% =========================================================================

\section{Implementation Details}
\label{app:implementation}

\subsection{Datasets and Evaluation Metrics}
\label{app:datasets}

\noindent \textbf{Datasets.} We train VideoSEG-O3 using a hierarchical three-stage pipeline, progressively endowing the model with distinct capabilities through tailored datasets. The specific composition of datasets for each stage is detailed in Table~\ref{tab:training_stages}.

\begin{itemize}
    \item \textbf{Stage I: Supervised Fine-Tuning (SFT).}
    \begin{itemize}
        \item \textit{Objective:} To equip the model with core competencies—including Video QA, Image/Video Segmentation, and Temporal Index Understanding—while specifically aligning the latent representation of the \texttt{[SEG]} token with the SAM decoder. Deviating from conventional timestamp-based methods, we explicitly employ frame-level \textbf{temporal indices} for temporal grounding. This design serves as a crucial foundation for Stage II and III, empowering the model to navigate frame sequences and develop a robust understanding of discrete temporal relationships.
        \item \textit{Datasets:}
        \begin{itemize}
            \item \textbf{Video QA:} ChatUniVi~\cite{chatunivi} (100K).
            \item \textbf{Temporal Grounding:} TimeLens-100K~\cite{timelens} (100K), utilized to enforce fine-grained frame index understanding.
            \item \textbf{Image RES:} RefCOCO/+/g~\cite{refcoco, refcocog} (56K) and GCG~\cite{glamm} (214K).
            \item \textbf{Video RES:} Ref-YTVOS~\cite{seo2020urvos} (3.5K), ReVOS~\cite{visa} (1.7K), MeViS~\cite{mevis} (0.6K), Ref-SAV~\cite{sa2va} (37K), and Long-RVOS~\cite{longrvos} (2.2K).
        \end{itemize}
    \end{itemize}

    \item \textbf{Stage II: Chain-of-Thought Cold-Start.}
    \begin{itemize}
        \item \textit{Objective:} To elicit multi-step reasoning, enable tool usage, and standardize the format for multi-turn interactions.
        \item \textit{Datasets (VTS-CoT):} We constructed a proprietary dataset (\textbf{VTS-CoT}, 6K samples) via GPT-assisted labeling. Source clips were selected from three challenging benchmarks (ReVOS~\cite{visa}, MeViS~\cite{mevis}, Long-RVOS~\cite{longrvos}) to ensure diversity in reasoning complexity, motion expression, and temporal duration.
    \end{itemize}

    \item \textbf{Stage III: Reinforcement Learning (RL).}
    \begin{itemize}
        \item \textit{Objective:} To optimize the model for multi-turn interactions and enhance segmentation accuracy by guiding the exploration of discriminative information via reward mechanisms.
        \item \textit{Datasets:} Training is conducted on challenging scenarios from ReVOS, and MeViS.
    \end{itemize}
\end{itemize}

\noindent \textbf{Evaluation Datasets and Metrics.} We evaluate \textbf{VideoSEG-O3} across eight major benchmarks to verify its reasoning and generalization capabilities. For Referring Video Object Segmentation, we report results on \textbf{MeViS}~\cite{mevis}, \textbf{Ref-Youtube-VOS}~\cite{seo2020urvos}, \textbf{Ref-DAVIS17}~\cite{ref-davis}, \textbf{Ref-SAV}~\cite{sa2va}, and \textbf{Long-RVOS}~\cite{longrvos}. For Reasoning Video Object Segmentation, we conduct evaluations on \textbf{ReVOS}~\cite{visa} (In-Domain), alongside zero-shot benchmarks on \textbf{ReasonVOS}~\cite{videolisa} and \textbf{GroundMoRe}~\cite{groundmore} (Out-Domain). Performance is measured using standard metrics: $\mathcal{J}$ (average Intersection over Union, IoU), $\mathcal{F}$ (boundary F-measure), and $\mathcal{J}\&\mathcal{F}$ (the average of $\mathcal{J}$ and $\mathcal{F}$).

\begin{table*}[h]
\centering
\caption{The three-stage training pipeline of VideoSEG-O3, detailing the specific capabilities developed and the datasets utilized at each stage.}
\label{tab:training_stages}
\small
\renewcommand{\arraystretch}{1.2} 
\begin{tabular}{llp{8.5cm}}
\toprule
\textbf{Stage} & \textbf{Capability} & \textbf{Training Datasets} \\
\midrule
\textbf{Stage I: SFT} & 
\begin{tabular}[t]{@{}l@{}} 
    Video QA \\ 
    Image/Video SEG \\ 
    Temporal Indices Understanding
\end{tabular} & 
\textbf{Video QA:} ChatUniVi~\cite{chatunivi} (100K) \newline
\textbf{Temporal Grounding:} TimeLens-100K~\cite{timelens} (100K) \newline
\textbf{Image RES:} RefCOCO/+/g~\cite{refcoco} (56K), ReasonSeg (0.2K), GCG (214K) \newline
\textbf{Video RES:} Ref-YTVOS~\cite{seo2020urvos} (3.5K), ReVOS~\cite{visa} (1.7K), MeViS~\cite{mevis} (0.6K), Ref-SAV~\cite{sa2va} (37K), Long-RVOS~\cite{longrvos} (2.2K) \\
\midrule
\textbf{Stage II: CoT Cold-Start} & 
\begin{tabular}[t]{@{}l@{}} 
    Step-wise Reasoning \\ 
    Tool Usage
\end{tabular} & 
\textbf{VTS-CoT (6K, Ours):} Curated from ReVOS~\cite{visa}, Long-RVOS~\cite{longrvos}, and MeViS~\cite{mevis} via GPT-assisted labeling. \\
\midrule
\textbf{Stage III: GRPO-RL} & 
\begin{tabular}[t]{@{}l@{}} 
    Multi-turn Interaction \\ 
    Mask Refinement
\end{tabular} & 
\textbf{Video RES:} MeViS~\cite{mevis} (0.6K), ReVOS~\cite{visa} (1.7K)\\
\bottomrule
\end{tabular}
\end{table*}

\subsection{Training Details}
\label{app:training}

The training of VideoSEG-O3 is hierarchically structured into three stages: supervised fine-tuning (SFT), Chain-of-Thought (CoT) cold-start, and Reinforcement Learning (RL). This section provides the exhaustive hyperparameter configurations for each phase.

\subsubsection{Stage I (SFT) and Stage II (CoT Cold-Start)}
The transition from Stage I to Stage II marks a significant shift from static image-level alignment to dynamic multi-frame temporal reasoning. While Stage I focuses on the fundamental modality bridge using static samples, Stage II introduces a decoupled visual sampling strategy to support the multi-turn Chain-of-Thought process. The detailed hyperparameter and visual input configurations are summarized in Table~\ref{tab:sft-cot-settings}.

\noindent \textbf{Decoupled Visual Input.} A pivotal difference in Stage II is the hierarchical processing of visual information. Specifically, we introduce up to 20 \textbf{temporal frames} sampled at a low resolution ($32 \times 28 \times 28$ max pixels) to provide global motion context without exceeding the MLLM's token limit. Simultaneously, a set of \textbf{spatial frames} (up to 5) is provided at a moderate resolution ($128 \times 28 \times 28$ max pixels) to capture environmental details. Crucially, for each reasoning turn, a single \textbf{keyframe} is injected at the highest resolution ($512 \times 28 \times 28$ max pixels).

\begin{table}[h]
\centering
\caption{Experimental settings and visual input configurations for Stage I (SFT) and Stage II (CoT Cold-start).}
\label{tab:sft-cot-settings}
\begin{tabular}{l|cc}
\toprule
\textbf{Configuration} & \textbf{Stage I (SFT)} & \textbf{Stage II (CoT Cold-start)} \\
\midrule
LoRA Rank ($r$) & 128 & 128 \\
LoRA Alpha ($\alpha$) & 256 & 256 \\
Backbone Frozen & Visual Encoder, LLM & Visual Encoder, LLM \\
Trainable Modules & SAM2 Decoder & SAM2 Decoder \\
\midrule
\textit{Visual Sampling Strategy} & & \\
Temporal Frames (Max) & None & 20 \\
Temporal Max Pixels & N/A & $32 \times 28 \times 28$ \\
Spatial Frames (Max) & 1 & 5 \\
Spatial Max Pixels & $512 \times 28 \times 28$ & $128 \times 28 \times 28$ \\
Keyframe per Round & N/A & 1 \\
Keyframe Max Pixels & N/A & $512 \times 28 \times 28$ \\
\midrule
Peak Learning Rate & $4 \times 10^{-5}$ & $4 \times 10^{-5}$ \\
Warmup Ratio & 0.05 & 0.05 \\
Max Sequence Length & 8192 & 8192 \\
\bottomrule
\end{tabular}
\end{table}

\subsubsection{Stage III: Reinforcement Learning (GRPO)}
Stage III leverages Group Relative Policy Optimization (GRPO)~\cite{deepseekr1} to facilitate autonomous exploration and refine the model's decision-making policy. Unlike previous stages, the RL phase employs full-parameter fine-tuning (excluding the frozen visual encoder) to maximize the expressive capacity of the policy. The detailed hyperparameter and visual configurations are summarized in Table~\ref{tab:rl-settings}.

\noindent \textbf{Dynamic Visual State in RL.} During the RL stage, the visual observation space $\mathcal{V}_t$ is updated dynamically to balance reasoning depth with computational efficiency. While spatial frame and keyframe resolutions remain consistent with the CoT stage, the temporal window is more strictly managed. We limit the total temporal frames to 10, with up to 5 newly sampled temporal frames added in each interaction round based on exploration actions. This ensures the model learns to prioritize informative temporal intervals within a compact context window.

\begin{table}[h]
\centering
\caption{Reinforcement Learning (Stage III) experimental settings. This stage prioritizes efficient multi-turn exploration through constrained temporal windows and high-resolution spatial anchors.}
\label{tab:rl-settings}
\begin{tabular}{l|l}
\toprule
\textbf{RL Configuration} & \textbf{Value} \\
\midrule
Algorithm & GRPO \\
Trainable Modules & LLM, SAM2 Decoder  \\
Optimization Policy & Full-parameter Fine-tuning \\
Learning Rate & $3 \times 10^{-6}$ \\
KL Coefficient ($\beta$) & 0.04 \\
Number of Generations ($G$) & 4 \\
Max Training Steps & 1000 \\
Max Interaction Turns & 3 \\
\midrule
\textit{Visual Sampling (RL-specific)} & \\
Initial Temporal Frames (Max) & 10 \\
New Temporal Frames per Round (Max) & 5 \\
Spatial Frames (Max) & 5 \\
\midrule
\textit{Resolution Constraints} & \\
Temporal Max Pixels & $32 \times 28 \times 28$ \\
Spatial Max Pixels & $128 \times 28 \times 28$ \\
Keyframe Max Pixels & $512 \times 28 \times 28$ \\
\midrule
Auxiliary Loss Ratio ($\lambda_{seg}$) & 0.2 \\
Numerical Precision & \texttt{bfloat16} \\
\bottomrule
\end{tabular}
\end{table}

\subsection{Inference Strategy}
\label{app:inference}

The inference phase maintains the multi-turn reasoning logic established in the RL stage but employs enhanced visual sampling strategies to optimize zero-shot transfer performance. Detailed configuration is provided in Table~\ref{tab:inference-settings}, with key technical enhancements highlighted below:

\begin{itemize}
    \item \textbf{Spatial Sampling Density}: To improve environment perception, we increase the number of \textbf{spatial frames} to 8, compared to the 5 frames used during RL training.
    \item \textbf{Resolution Scaling}: While the temporal and keyframe resolutions remain consistent with Stage III, the \textbf{spatial resolution} is enhanced to $256 \times 28 \times 28$ pixels (doubling the training resolution) to provide clearer semantic cues for object localization.
    \item \textbf{Execution Constraints}: The system is constrained to a maximum of 3 interaction turns, processing up to 20 sampled video frames to ensure a comprehensive temporal horizon.
\end{itemize}

\begin{table}[ht]
\centering
\caption{Inference hyperparameter configurations for VideoSEG-O3. Values in parentheses denote the increments or scaling factors relative to the Reinforcement Learning training stage (Stage III).}
\label{tab:inference-settings}
\begin{tabular}{l|l}
\toprule
\textbf{Inference Parameter} & \textbf{Value} \\
\midrule
Max Interaction Turns & 3 \\
Max Video Sample Frames & 20 \\
Spatial Sampling Frames ($K$) & 8 \quad (\textbf{+3 vs. RL}) \\
\midrule
\textit{Resolution Constraints (Max Pixels)} & \\
Temporal Resolution & $32 \times 28 \times 28$ \\
Spatial Resolution & $256 \times 28 \times 28$ \quad (\textbf{$\times$2 vs. RL}) \\
Keyframe Resolution & $512 \times 28 \times 28$ \\
\bottomrule
\end{tabular}
\end{table}

% =========================================================================
% Section B: VTS-CoT Construction
% =========================================================================
\clearpage
\section{VTS-CoT Construction Details}
\label{app:vtscot}

\subsection{Pipeline Overview}
\label{app:vtscot_pipeline}

\begin{figure*}[h]
    \centering
    \includegraphics[width=1.0\textwidth]{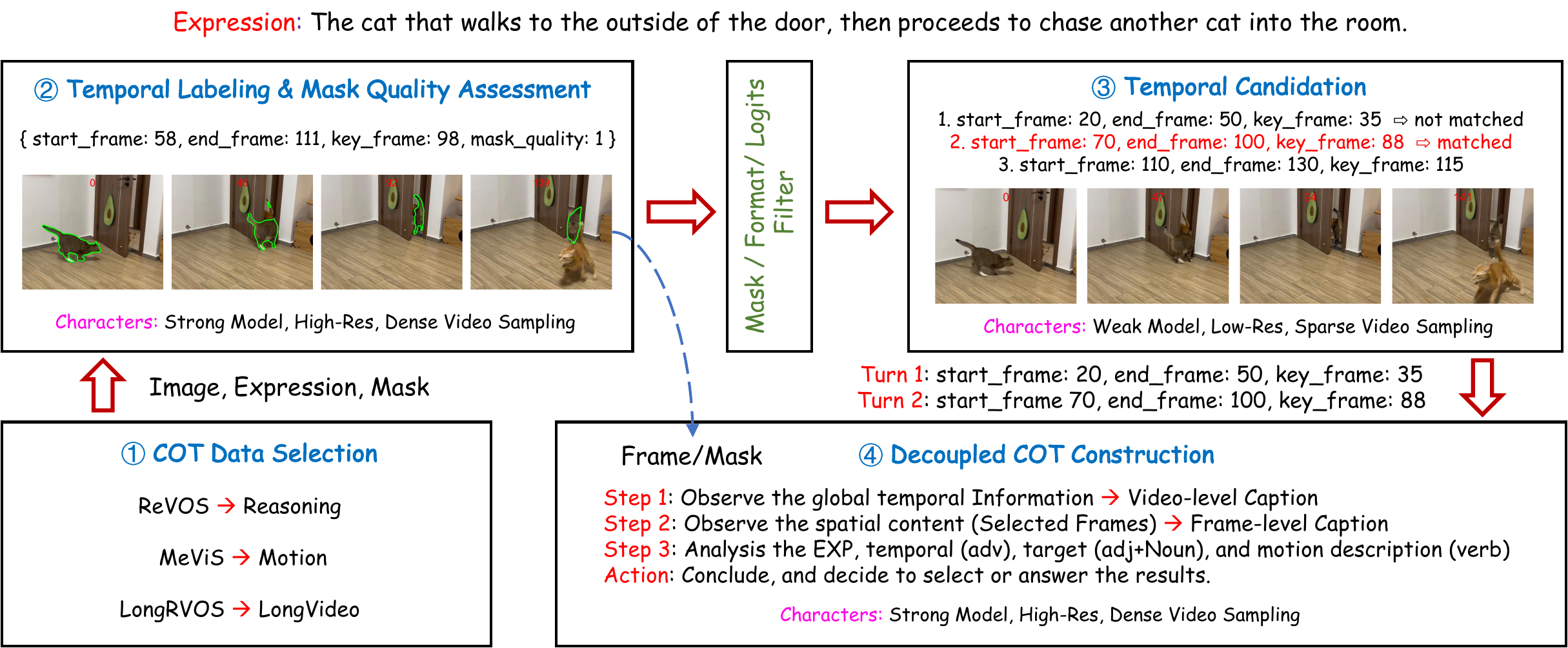} 
    \caption{Pipeline of our VTS-CoT: Data selection $\rightarrow$ Temporal labeling $\rightarrow$ Candidate generation $\rightarrow$ Chain-of-Thought construction.}
    \label{fig:cotpipeline}
\end{figure*}

As illustrated in Fig.~\ref{fig:cotpipeline}, the construction pipeline of VTS-CoT is structured into four distinct components: data curation, temporal labeling, candidate generation, and Chain-of-Thought (CoT) synthesis. In this section, we elaborate on the design rationale and implementation details for each component.

\noindent \textbf{1. Diverse Data Curation.} 
We aggregate a diverse collection of video clips from three representative benchmarks: ReVOS, Long-RVOS, and MeViS. Each dataset introduces unique challenges to ensure the complexity and generalization capability of VTS-CoT. Specifically, ReVOS focuses on reasoning-intensive scenarios, MeViS emphasizes complex motion and expression understanding, while Long-RVOS contributes samples characterized by extended temporal durations.

\noindent \textbf{2. Temporal Labeling and Mask Quality Assessment.} 
We design a specialized prompt (see Table~\ref{tab:vts-cot-step2-prompt}) to guide the Qwen3VL-235B-A30B-Instruct~\cite{Qwen3-VL} model in performing precise temporal labeling and evaluating mask fidelity. Following the initial annotation, we implement a rigorous filtering protocol to retain high-quality samples. To ensure annotation accuracy, we employ a logical dual-verification mechanism that cross-references the temporal grounding results with the ground-truth mask targets; samples exhibiting significant discrepancies are discarded. Furthermore, to enhance the model's temporal grounding precision, we adopt a high-resolution, dense sampling strategy and explicitly inject mask information into the prompt to serve as visual auxiliary cues.

\noindent \textbf{3. Temporal Candidation.} 
Leveraging the refined temporal annotations from the previous stage, we generate three top-ranked candidate temporal intervals. This process is executed by the ERNIE-45-VL-424B-A47B model using the prompt detailed in Table~\ref{tab:vts-cot-step3-prompt}. To encourage the generation of diverse yet plausible temporal proposals, we deliberately employ a low-resolution, sparse video sampling strategy during this phase.

\noindent \textbf{4. Decoupled CoT Construction.} 
With the high-quality temporal annotations and candidate intervals established, we utilize the prompt presented in Table~\ref{tab:vts-cot-step4-prompt} to instruct the Qwen3VL-235B-A30B-Thinking~\cite{Qwen3-VL} model in synthesizing the final reasoning chains. To maximize the diversity and coverage of the generated Chain-of-Thought (CoT) data, we apply a hybrid video sampling strategy, alternating between high-resolution dense sampling and low-resolution sparse sampling to simulate varying observational contexts.

\subsection{Prompt Designs}
\label{app:vtscot_prompts}

We provide the prompt templates used for the distinct stages of VTS-CoT construction below.

\begin{table*}[ht]
  \centering
  \begin{tcolorbox}[
    enhanced,
    width=\linewidth,
    colback=bg_gray,
    colframe=border_gray,
    boxrule=0.8pt,
    arc=2mm,
    left=1em, right=1em, top=1em, bottom=1em,
    fonttitle=\bfseries\sffamily,
    title={Prompt for Temporal Labeling and Mask Quality Assessment},
    coltitle=black,
    attach boxed title to top left={xshift=1.5em, yshift*=-\tcboxedtitleheight/2},
    boxed title style={colback=white, boxrule=0.8pt, colframe=border_gray}
  ]
    \small\sffamily
    
    % --- Role ---
    \textbf{Role \& Objective:}
    \par\noindent
    Act as a specialist in Spatiotemporal Video Grounding and Annotation Verification. Your objective is to precisely localize a specific event temporally and evaluate the fidelity of pre-annotated spatial segmentation masks.
    
    \vspace{0.6em}
    \hrule \vspace{0.6em}
    
    % --- Inputs ---
    \textbf{Input Data:}
    \begin{itemize}[leftmargin=1.2em, label={\textcolor{gray}{\small$\bullet$}}, noitemsep, topsep=0pt]
        \item \textbf{Target Query:} \texttt{<referring\_expression>}
        \item \textbf{Sampled Frames:} A uniform sequence of video frames containing:
        \begin{itemize}[leftmargin=1em, label=-]
           \item \textit{Visual Indices:} \textcolor{red}{Red numbers} indicating the temporal index of each frame.
           \item \textit{Candidate Masks:} \textcolor{green}{Green contour} outlines representing the ground truth proposals.
        \end{itemize}
        \item \textbf{Video Metadata:} Total frame count:\texttt{<total\_frames>}, denoted as $T_{total}$.
    \end{itemize}

    \vspace{0.6em}
    \hrule \vspace{0.6em}

    \textbf{Tasks:}
    \begin{enumerate}[leftmargin=1.2em, label=\textbf{\arabic*.}, noitemsep, topsep=0pt]
        \item \textbf{Temporal Localization:} Identify the most semantic-relevant time interval $[t_{start}, t_{end}]$ and a representative keyframe $t_{key}$ that best aligns with the \textit{Target Query}.
        \item \textbf{Mask Fidelity Assessment:} Evaluate the \textit{Candidate Masks}. Determine if the green contours accurately and consistently capture the target object described in the query.
    \end{enumerate}

    \vspace{0.6em}
    \hrule \vspace{0.6em}

    \textbf{Constraints \& Rules:}
    \begin{itemize}[leftmargin=1.2em, label={\textcolor{gray}{\small$\bullet$}}, noitemsep, topsep=0pt]
        \item \textbf{Temporal Duration:} The selected window must capture the core event without being excessive. The duration $\Delta t = t_{end} - t_{start}$ must satisfy:
        \[ 5 < \Delta t < 0.5 \times T_{total} \]
        \item \textbf{Index Validity:} Ensure logical consistency of frame indices:
        \[ 0 \le t_{start} \le t_{key} \le t_{end} < T_{total} \]
        \item \textbf{Quality Criteria:}
        \begin{itemize}[leftmargin=1em, label=$\cdot$]
            \item \textbf{1 (Valid):} The mask is complete, temporally consistent, and semantically correct.
            \item \textbf{0 (Invalid):} The mask is fragmented, drifting, absent, or misaligned with the target.
        \end{itemize}
    \end{itemize}

    \vspace{0.6em}
    \hrule \vspace{0.6em}

    % --- Output ---
    \textbf{Output Schema:}
    \par\noindent
    Return strictly a single JSON object with the following structure:
    
    \vspace{0.4em}
    \begin{tcolorbox}[colback=white, colframe=gray!30, boxrule=0.5pt, sharp corners, left=2mm, top=1mm, bottom=1mm]
    \ttfamily\textcolor{code_blue}{\{"start\_frame": <int>, "end\_frame": <int>, "key\_frame": <int>, "mask\_quality": <0|1>\}}
    \end{tcolorbox}

  \end{tcolorbox}
  \vspace{-0.5em}
  \caption{The structured prompt template used for GPT-assisted dataset curation. We formalize the constraints using mathematical inequalities to enforce precise temporal localization.}
  \label{tab:vts-cot-step2-prompt}
\end{table*}

\begin{table*}[ht]
  \centering
  \begin{tcolorbox}[
    enhanced,
    width=\linewidth,
    colback=bg_gray,
    colframe=border_gray,
    boxrule=0.8pt,
    arc=2mm,
    left=1em, right=1em, top=1em, bottom=1em,
    fonttitle=\bfseries\sffamily,
    title={Prompt for Top-3 Non-Overlapping Temporal Candidation},
    coltitle=black,
    attach boxed title to top left={xshift=1.5em, yshift*=-\tcboxedtitleheight/2},
    boxed title style={colback=white, boxrule=0.8pt, colframe=border_gray}
  ]
    \small\sffamily
    
    % --- Role ---
    \textbf{Role \& Objective:}
    \par\noindent
    Act as a specialist in Multi-Event Video Analysis. Your task is to identify and rank the Top-3 most relevant, non-overlapping temporal segments within a video that align with a specific text query.
    
    \vspace{0.6em}
    \hrule \vspace{0.6em}
    
    % --- Inputs ---
    \textbf{Input Data:}
    \begin{itemize}[leftmargin=1.2em, label={\textcolor{gray}{\small$\bullet$}}, noitemsep, topsep=0pt]
        \item \textbf{Target Query:} \texttt{<referring\_expression>}
        \item \textbf{Visual Context:} Sampled video frames with visual indices (red numbers) indicating temporal location.
        \item \textbf{Metadata:} Total frame count:\texttt{<total\_frames>}, denoted as $T_{total}$.
    \end{itemize}

    \vspace{0.6em}
    \hrule \vspace{0.6em}

    % --- Strategy (核心部分) ---
    \textbf{Iterative Exclusion Strategy:}
    \par\noindent
    To ensure diversity and strict ranking, follow this sequential selection process:
    \begin{enumerate}[leftmargin=1.2em, label=\textbf{\arabic*.}, noitemsep, topsep=2pt]
        \item \textbf{Primary Selection ($S_1$):} Identify the single most relevant segment in the entire video.
        \item \textbf{Secondary Selection ($S_2$):} Mask the time range of $S_1$. Find the best matching segment in the remaining video parts ($V \setminus S_1$).
        \item \textbf{Tertiary Selection ($S_3$):} Mask the time ranges of both $S_1$ and $S_2$. Find the best match in ($V \setminus (S_1 \cup S_2)$).
    \end{enumerate}

    \vspace{0.6em}
    \hrule \vspace{0.6em}

    % --- Constraints (公式化) ---
    \textbf{Constraints \& Rules:}
    \begin{itemize}[leftmargin=1.2em, label={\textcolor{gray}{\small$\bullet$}}, noitemsep, topsep=0pt]
        \item \textbf{Relevance Ranking:} Output must be sorted by semantic relevance: $Rel(S_1) > Rel(S_2) > Rel(S_3)$.
        \item \textbf{Non-Overlapping:} Segments must be mutually exclusive:
        \[ S_i \cap S_j = \emptyset, \quad \forall i \neq j \]
        \item \textbf{Duration Constraints:} For any segment $S_i$ with duration $\Delta t_i$:
        \[ 0.1 \times T_{total} \le \Delta t_i < 0.33 \times T_{total} \]
        \item \textbf{Index Validity:} $0 \le t_{start} \le t_{key} \le t_{end} < T_{total}$.
        \item \textbf{Exception Handling:} Return fewer than 3 segments only if the video lacks sufficient distinct content.
    \end{itemize}

    \vspace{0.6em}
    \hrule \vspace{0.6em}

    % --- Output ---
    \textbf{Output Schema:}
    \par\noindent
    Output \textbf{ONLY} a raw JSON list. Do not use markdown blocks.
    
    \vspace{0.4em}
    \begin{tcolorbox}[colback=white, colframe=gray!30, boxrule=0.5pt, sharp corners, left=2mm, top=1mm, bottom=1mm]
    \ttfamily\textcolor{code_blue}{[
   \{"start\_frame": <int>, "end\_frame": <int>, "key\_frame": <int>\},
   \{"start\_frame": <int>, "end\_frame": <int>, "key\_frame": <int>\},
   ...
]}
    \end{tcolorbox}

  \end{tcolorbox}
  \vspace{-0.5em}
  \caption{The prompt template for the Top-3 iterative temporal grounding task. We explicitly enforce an exclusion strategy to generate diverse, non-overlapping proposals.}
  \label{tab:vts-cot-step3-prompt}
\end{table*}

\begin{table*}[ht]
  \centering
  \begin{tcolorbox}[
    enhanced,
    width=\linewidth,
    colback=bg_gray, % Ensure this color is defined in your preamble
    colframe=border_gray, % Ensure this color is defined in your preamble
    boxrule=0.8pt,
    arc=2mm,
    left=1em, right=1em, top=1em, bottom=1em,
    fonttitle=\bfseries\sffamily,
    title={Prompt for Sequential Spatiotemporal CoT Construction},
    coltitle=black,
    attach boxed title to top left={xshift=1.5em, yshift*=-\tcboxedtitleheight/2},
    boxed title style={colback=white, boxrule=0.8pt, colframe=border_gray}
  ]
    \small\sffamily
    
    % --- Role ---
    \textbf{Role \& Objective:}
    \par\noindent
    Act as an Advanced Spatiotemporal Reasoning Agent. Simulate a strictly sequential human reasoning process to locate a target described by a text expression. Generate a logical ``Chain of Thought'' (CoT) justifying segment selection, moving from global overview to specific verification.
    
    \vspace{0.6em}
    \hrule \vspace{0.6em}
    
    % --- Task Simulation ---
    \textbf{Task Simulation (Two Phases):}
    \begin{enumerate}[leftmargin=1.2em, label=\textbf{\arabic*.}, noitemsep, topsep=0pt]
        \item \textbf{Phase 1 (Initialization):} Analyze global context and text to form a hypothesis. \textbf{Constraint:} You cannot access specific segment details yet. \textbf{Action:} Request Segment 0.
        \item \textbf{Phase 2 (Verification Loop):} Analyze specific frames requested in the previous step. Verify if the target matches. \textbf{Action:} Request next segment OR Confirm final target.
    \end{enumerate}

    \vspace{0.6em}
    \hrule \vspace{0.6em}

    % --- Inputs ---
    \textbf{Input Data:}
    \begin{itemize}[leftmargin=1.2em, label={\textcolor{gray}{\small$\bullet$}}, noitemsep, topsep=0pt]
        \item \textbf{Global:} Referring Expression, Total Video Length, Sampled Frame Indices.
        \item \textbf{Input Segments:} Sequence of temporal segments (start/end/keyframe).
        \item \textbf{Visual Cues:} \textcolor{red}{Red Numbers} (Frame Indices), \textcolor{green}{Green Contours} (Target location, internal use only).
    \end{itemize}

    \vspace{0.6em}
    \hrule \vspace{0.6em}

    % --- Reasoning Workflow ---
    \textbf{Strict Reasoning Workflow (Output JSON list of size $N+1$):}
    \begin{enumerate}[leftmargin=1.2em, label=\textbf{\arabic*.}, noitemsep, topsep=2pt]
        \item \textbf{First Item (Global Analysis):}
        \begin{itemize}[leftmargin=1em, label=-, noitemsep]
            \item \textit{Step 1 (Global Temporal):} Analyze low-res global context for candidates.
            \item \textit{Step 2 (Local Spatial):} Inspect Sampled Frames.
            \item \textit{Step 3 (Alignment):} Correlate visual cues with text constraints.
            \item \textit{Action:} Request \texttt{Input Segment [0]}.
        \end{itemize}
        \item \textbf{Subsequent Items (Verification Loop for Segment $k$):}
        \begin{itemize}[leftmargin=1em, label=-, noitemsep]
            \item \textit{Step 1 (Segment Temporal):} Analyze motion in the requested interval.
            \item \textit{Step 2 (Segment Spatial):} Verify visual attributes in the requested \texttt{key\_frame}.
            \item \textit{Step 3 (Refinement):} Compare specific evidence against text.
            \item \textit{Action:} Justify checking \texttt{Segment [k+1]} OR Confirm \texttt{Final Segment}.
        \end{itemize}
    \end{enumerate}

    \vspace{0.6em}
    \hrule \vspace{0.6em}

    % --- Rules ---
    \textbf{Constraints \& Rules:}
    \begin{itemize}[leftmargin=1.2em, label={\textcolor{gray}{\small$\bullet$}}, noitemsep, topsep=0pt]
        \item \textbf{Information Isolation:} You only see data requested in the \textit{previous} Action. No peeking ahead.
        \item \textbf{No Visual Leakage:} Never mention "green contours" or "outlines" in the output. Describe the object itself.
        \item \textbf{Format:} Single-line, valid raw JSON list. No markdown.
    \end{itemize}

    \vspace{0.6em}
    \hrule \vspace{0.6em}

    % --- Few Shot (Condensed) ---
    \textbf{Few-Shot Strategy (Example: Single Segment):}
    \par\noindent
    \textit{Exp:} ``The car at the very front at the beginning...''
    \vspace{0.2em}
    \begin{tcolorbox}[colback=white, colframe=gray!30, boxrule=0.5pt, sharp corners, left=2mm, top=1mm, bottom=1mm]
    \ttfamily\scriptsize\textcolor{code_blue}{[
    \{"think": "Step 1: Global view shows a race... Step 2: At frame 0, red car leads... Step 3: 'beginning' requires focusing on leader. Action: Inspect 0-23 segment...", "start\_frame": 0, "end\_frame": 23, "key\_frame": 5\}, 
    \{"think": "Step 1: Segment 0-23 shows yellow Mercedes... Step 2: Keyframe 4 shows it at front... Step 3: Matches 'very front'. Action: Target confirmed.", "start\_frame": 0, "end\_frame": 10, "key\_frame": 4\}
]}
    \end{tcolorbox}

    \vspace{0.6em}
    \hrule \vspace{0.6em}

    % --- Output ---
    \textbf{Output Schema:}
    \par\noindent
    Output \textbf{ONLY} a raw JSON list.
    \vspace{0.4em}
    \begin{tcolorbox}[colback=white, colframe=gray!30, boxrule=0.5pt, sharp corners, left=2mm, top=1mm, bottom=1mm]
    \ttfamily\scriptsize\textcolor{code_blue}{[
  \{"think": "Step 1: [Temporal]. Step 2: [Spatial]. Step 3: [Expression]. Action: [Decision].", "start\_frame": <int>, "end\_frame": <int>, "key\_frame": <int>\}, ...
]}
    \end{tcolorbox}

  \end{tcolorbox}
  \vspace{-0.5em}
  \caption{The system prompt for the Spatiotemporal Reasoning Agent. It enforces a strict, two-phase CoT workflow (Global Analysis $\to$ Sequential Verification) with information isolation to simulate human-like video grounding.}
  \label{tab:vts-cot-step4-prompt}
\end{table*}

\clearpage

\section{Additional Methods}
\label{app:methods}
\subsection{VideoSEG-O3 Reward Design Details}
\label{app:reward_design}

In this section, we provide the mathematical formulation for the composite reward function $\mathcal{R}$, which consists of four components: Format Reward ($\mathcal{R}_f$), Temporal Reward ($\mathcal{R}_t$), Segmentation Quality Reward ($\mathcal{R}_m$), and Progressive Reward ($\mathcal{R}_p$).

\subsubsection{Format Reward ($\mathcal{R}_f$)}
To ensure the model adheres to the strict protocols required for parsing and execution, we implement a two-stage format check.
\begin{itemize}
    \item \textbf{Think Format:} We parse the \texttt{<think>} block to verify the existence of the Chain-of-Thought structure. The reward is accumulated based on the presence of four required keys: ``Step 1'', ``Step 2'', ``Step 3'', and ``Action''. Each successfully detected key yields $+0.25$, summing to a maximum of $1.0$.
    \item \textbf{Result Format:} We validate the syntactic correctness of the final answer block. The output must strictly follow the format: a \texttt{<VTG>} dictionary containing \texttt{start}, \texttt{end}, and \texttt{keyframe} fields, followed by a \texttt{<RefVOS>} tag enclosing the \texttt{[SEG]} token. If the output cannot be parsed or the intervals are invalid ($start \ge end$), a penalty (e.g., $-0.5$) is applied; otherwise, it is eligible for subsequent dense rewards.
\end{itemize}
\subsubsection{Temporal Reward ($\mathcal{R}_t$)}
The temporal reward evaluates the model's grounding capability from two perspectives: discrete keyframe selection and continuous temporal range generation. Let $E_m \in \{0, 1\}^{T}$ be the ground-truth binary sequence and $[S, E]$ be the predicted temporal interval.

\paragraph{Keyframe Alignment Reward ($\mathcal{R}_t^k$):}
This component validates whether the selected keyframe index $k$ falls within the valid action range. It acts as a strict binary feedback:
\begin{equation}
    \mathcal{R}_t^k = 
    \begin{cases} 
    +1.0 & \text{if } E_m[k] = 1, \\
    -1.0 & \text{if } E_m[k] = 0.
    \end{cases}
\end{equation}

\paragraph{Temporal Precision Reward ($\mathcal{R}_t^p$):}
To evaluate the quality of the predicted interval $[S, E]$, we calculate the temporal precision $P_t$, defined as the proportion of the predicted frames that actually contain the target: $P_t = \frac{\sum_{t=S}^{E} E_m[t]}{E - S}$. The dense reward is then calculated as:
\begin{equation}
    \mathcal{R}_t^p = 
    \begin{cases} 
    0.5 \times P_t & \text{if } P_t \ge 0.5, \\
    -0.5 & \text{if } P_t < 0.5.
    \end{cases}
\end{equation}
This formulation penalizes loose predictions (low precision) with a fixed penalty of $-0.5$, while scaling the positive reward linearly with precision when it exceeds the $0.5$ threshold.

\subsubsection{Segmentation Quality Reward ($\mathcal{R}_m$)}
To stabilize training, we map continuous IoU values to discrete reward tiers using piecewise functions.

\paragraph{Spatial Reward ($\mathcal{R}_m^s$):}
This rewards the average mask quality across sampled spatial frames. Let $\text{IoU}_s$ be the average IoU of the predicted masks in the final turn. The mapping function is defined as:
\begin{equation}
    \mathcal{R}_m^s = 
    \begin{cases} 
    \min(3(\text{IoU}_s - 0.4), 1.5) & \text{if } \text{IoU}_s > 0.4, \\
    0.0 & \text{if } 0.2 < \text{IoU}_s \le 0.4, \\
    -1.0 & \text{if } \text{IoU}_s \le 0.2.
    \end{cases}
\end{equation}

\paragraph{Keyframe Superiority Reward ($\mathcal{R}_m^k$):}
This component incentivizes the model to identify a keyframe that offers a significantly better view than the average spatial context. It is based on the differential $\Delta_{k} = \text{IoU}_k - \text{IoU}_s$:
\begin{equation}
    \mathcal{R}_m^k = 
    \begin{cases} 
    \min(5(\Delta_{k} - 0.1), 2.0) & \text{if } \Delta_{k} > 0.2, \\
    0.5 & \text{if } 0.1 < \Delta_{k} \le 0.2, \\
    0.0 & \text{if } -0.05 < \Delta_{k} \le 0.1, \\
    -1.0 & \text{if } \Delta_{k} \le -0.05.
    \end{cases}
\end{equation}

\subsubsection{Progressive Reward ($\mathcal{R}_p$)}
To encourage continuous improvement across multi-turn interactions, we compare the current keyframe IoU ($\text{IoU}_{k,t}$) with the best performance from previous turns ($\text{IoU}_{\text{best}} = \max_{j<t} \text{IoU}_{k,j}$). Let $\Delta_p = \text{IoU}_{k,t} - \text{IoU}_{\text{best}}$. The progressive reward is defined as:
\begin{equation}
    \mathcal{R}_p = 
    \begin{cases} 
    \min(10(\Delta_p - 0.05), 2.0) & \text{if } \Delta_p > 0.1, \\
    0.5 & \text{if } 0.05 < \Delta_p \le 0.1, \\
    0.0 & \text{if } -0.05 < \Delta_p \le 0.05, \\
    -1.0 & \text{if } \Delta_p \le -0.05.
    \end{cases}
\end{equation}
This mechanism strictly penalizes degradation ($\Delta_p \le -0.05$) and offers substantial rewards (slope of 10) only when the model breaks previous records by a significant margin ($\Delta_p > 0.1$).
% =========================================================================
% Section C: Additional Results
% =========================================================================
\clearpage
\section{Additional Quantitative Results}
\label{app:results}

\subsection{Results on GroundMoRe Tasks}
\label{app:groundmore}
GroundMoRe~\cite{groundmore} is a challenging benchmark requiring joint spatio-temporal reasoning for comprehensive video understanding. To evaluate the generalization capability of VideoSEG-O3, we conducted zero-shot evaluations on this dataset. The results, summarized in Table~\ref{tab:groundmore_results}, demonstrate that VideoSEG-O3 achieves a new state-of-the-art with an Overall $\mathcal{J}\&\mathcal{F}$ score of \textbf{31.96\%}, significantly outperforming the previous best method, MORA, by \textbf{+8.83\%}.
Specifically, our model yields remarkable improvements in the \textit{Descriptive} (+14.76\%) and \textit{Causal} (+13.06\%) categories. This highlights the robustness of our Multi-turn CoT in capturing fine-grained spatio-temporal attributes and tracking complex cause-effect chains.
However, we observe a slight performance dip in the \textit{Counterfactual} category (-1.31\% compared to MORA). We attribute this primarily to the training data distribution gap inherent in the zero-shot setting.

\begin{table*}[ht]
\centering
\vspace{-2mm}
\caption{Comparison of Motion-Grounded Video Reasoning results on our benchmark in a zero-shot setting. Best results are highlighted in \textbf{bold}.}
\vspace{-2mm}
\resizebox{1.0\textwidth}{!}{
\begin{tabular}{l|ccc|ccc|ccc|ccc|ccc}
\toprule
\multirow{2}{*}{\textbf{Methods}} & \multicolumn{3}{c}{\textbf{Overall}} & \multicolumn{3}{c}{\textbf{Causal}} & \multicolumn{3}{c}{\textbf{Sequential}} & \multicolumn{3}{c}{\textbf{Counterfactual}} & \multicolumn{3}{c}{\textbf{Descriptive}} \\ 
\cmidrule(lr){2-4} \cmidrule(lr){5-7} \cmidrule(lr){8-10} \cmidrule(lr){11-13} \cmidrule(lr){14-16}
 & \textbf{J\&F} & \textbf{J} & \textbf{F} & \textbf{J\&F} & \textbf{J} & \textbf{F} & \textbf{J\&F} & \textbf{J} & \textbf{F} & \textbf{J\&F} & \textbf{J} & \textbf{F} & \textbf{J\&F} & \textbf{J} & \textbf{F} \\ 
\midrule
\multicolumn{16}{l}{\textbf{\textit{Random Baseline}}} \\
Title+ReferFormer~\citep{wu2022referformer} & 10.13 & 9.73 & 10.53 & 10.50 & 9.91 & 11.08 & 9.06 & 8.56 & 9.57 & 9.42 & 9.07 & 9.78 & 11.38 & 11.21 & 11.55 \\ 
\midrule
\multicolumn{16}{l}{\textbf{\textit{RVOS Baseline}}} \\ 
ReferFormer~\citep{wu2022referformer} & 12.72 & 11.15 & 14.29 & 12.67 & 10.95 & 14.38 & 10.73 & 9.35 & 12.10 & 12.36 & 11.12 & 13.61 & 15.15 & 13.26 & 17.04 \\ 
SgMg~\citep{miao2023spectrum} & 17.49 & 15.81 & 19.16 & 18.94 & 17.04 & 20.85 & 15.46 & 13.97 & 16.96 & 16.35 & 15.24 & 17.45 & 18.77 & 16.69 & 20.86 \\ 
HTR~\citep{miao2024temporally} & 14.48 & 12.71 & 16.24 & 15.66 & 13.82 & 17.51 & 12.46 & 10.78 & 14.14 & 13.51 & 12.16 & 14.86 & 15.94 & 13.80 & 18.08 \\ 
LMPM~\citep{mevis} & 13.34 & 12.71 & 13.97 & 13.71 & 13.13 & 14.30 & 11.30 & 10.60 & 12.01 & 13.16 & 12.64 & 13.68 & 14.33 & 13.60 & 15.05 \\
\midrule
\multicolumn{16}{l}{\textbf{\textit{Image Reasoning Segmentation Baseline}}} \\ 
LISA-7B~\citep{lisa} & 6.11 & 4.89 & 7.32 & 5.53 & 4.38 & 6.69 & 5.19 & 4.07 & 6.32 & 5.88 & 4.77 & 6.98 & 7.56 & 6.11 & 9.00 \\ 
PixelLM-7B~\citep{pixellm} & 9.96 & 9.93 & 9.99 & 9.52 & 9.39 & 9.65 & 9.21 & 9.15 & 9.27 & 9.82 & 9.76 & 9.87 & 10.94 & 11.02 & 10.85 \\ 
\midrule
\multicolumn{16}{l}{\textbf{\textit{Video Reasoning Segmentation Baseline}}} \\ 
PG-Video-LLaVA~\citep{munasinghe2023pg} & 11.17 & 11.47 & 10.89 & 10.57 & 10.82 & 10.33 & 10.93 & 11.15 & 10.71 & 10.36 & 10.37 & 10.34 & 12.51 & 13.04 & 11.99 \\ 
PG-Video-LLaVA+SAM2~\citep{sam2} & 11.85 & 11.15 & 12.55 & 13.35 & 12.84 & 13.85 & 3.85 & 2.49 & 5.22 & 15.52 & 15.16 & 15.89 & 12.11 & 11.37 & 12.85 \\ 
VISA~\citep{visa} & 5.31 & 4.72 & 5.90 & 6.07 & 5.50 & 6.64 & 4.48 & 3.80 & 5.15 & 5.10 & 4.34 & 5.86 & 5.39 & 5.02 & 5.77 \\ 
\midrule
\multicolumn{16}{l}{\textbf{\textit{Two-Stage Baseline}}} \\ 
ViLA~\citep{vila}+ReferFormer & 16.37 & 14.84 & 17.89 & 15.92 & 14.33 & 17.52 & 16.30 & 14.94 & 17.67 & 13.63 & 12.35 & 14.91 & 19.61 & 17.78 & 21.45 \\ 
ViLA~\citep{vila}+SgMg & 17.07 & 15.25 & 18.89 & 17.34 & 15.43 & 19.26 & 16.30 & 14.57 & 18.04 & 15.31 & 13.96 & 16.67 & 19.18 & 16.96 & 21.40 \\ 
ViLA~\citep{vila}+HTR & 15.72 & 13.90 & 17.53 & 16.46 & 14.60 & 18.31 & 15.11 & 13.39 & 16.83 & 12.61 & 11.22 & 13.99 & 18.37 & 16.10 & 20.63 \\ 
VideoChat2~\citep{mvbench}+ReferFormer & 15.38 & 13.84 & 16.93 & 14.74 & 13.10 & 16.37 & 13.95 & 12.69 & 15.21 & 13.12 & 11.80 & 14.43 & 19.82 & 17.89 & 21.74 \\ 
VideoChat2~\citep{mvbench}+SgMg & 16.65 & 14.83 & 18.48 & 16.64 & 14.75 & 18.54 & 14.70 & 13.02 & 16.38 & 15.22 & 13.87 & 16.58 & 20.01 & 17.68 & 22.34 \\ 
VideoChat2~\citep{mvbench}+HTR & 15.08 & 13.25 & 16.90 & 15.91 & 14.00 & 17.81 & 13.91 & 12.33 & 15.50 & 11.97 & 10.56 & 13.37 & 18.17 & 15.82 & 20.52 \\ 
SeViLA~\citep{yu2023self}+ReferFormer & 20.37 & 18.99 & 21.75 & 21.29 & 19.86 & 22.72 & 18.27 & 17.06 & 19.48 & 17.56 & 16.37 & 18.75 & 24.01 & 22.35 & 25.66 \\ 
SeViLA~\citep{yu2023self}+SgMg & 22.34 & 20.92 & 23.75 & 21.60 & 20.17 & 23.02 & 21.68 & 20.47 & 22.89 & 20.46 & 19.32 & 21.61 & 24.53 & 22.83 & 26.22 \\ 
SeViLA~\citep{yu2023self}+HTR & 21.75 & 19.85 & 23.64 & 22.28 & 20.42 & 24.15 & 20.37 & 18.60 & 22.14 & 19.70 & 18.03 & 21.36 & 24.41 & 22.11 & 26.71 \\ 
\midrule
MORA~\cite{groundmore} & 23.13 & 22.79 & 23.46 & 21.62 & 21.46 & 21.79 & 22.43 & 22.03 & 22.83 & \textbf{26.37} & \textbf{25.87} & \textbf{26.88} & 22.63 & 22.30 & 22.97 \\
\midrule
\textbf{VideoSEG-O3-2B (Ours)} & 29.06 & 27.39 & 30.72 & 30.93 & 29.21 & 32.65 & 26.97 & 25.42 & 28.52 & 22.04 & 20.62 & 23.46 &  35.49 & 33.54 & 37.44  \\
\textbf{VideoSEG-O3-4B (Ours)} & \textbf{31.96} & \textbf{30.19} & \textbf{33.73} & \textbf{34.68} & \textbf{32.88} & \textbf{36.47} & \textbf{29.71} & \textbf{28.14} & \textbf{31.28} & 25.06 & 23.51 & 26.60 & \textbf{37.39} & \textbf{35.21} & \textbf{39.56} \\
\bottomrule
\end{tabular}
}
\label{tab:groundmore_results}
\end{table*}

% =========================================================================
% Section D: Additional Ablation Studies
% =========================================================================
\section{Additional Ablation Studies}
\label{app:ablation}

\subsection{Effect of Post-Processing Strategies}
\label{app:post_process}
To investigate optimal inference strategies for leveraging multi-turn reasoning in high-fidelity mask generation, we evaluate three post-processing methods: \textit{Sa2VA-type First N}, \textit{Keyframe Bidirectional}, and \textit{Language Injection} (Table~\ref{tab:ablation_postprocess}). The significant performance gains observed with the \textit{Keyframe Bidirectional} strategy validate VideoSEG-O3's proficiency in its two primary optimization objectives: \textbf{Keyframe Localization} and \textbf{Embedding Refinement}. By successfully identifying the most discriminative keyframe where the target is salient, the model establishes a robust visual anchor for bidirectional temporal propagation, demonstrating that precise spatiotemporal understanding is critical for maximizing segmentation performance in multi-turn interactions.

\begin{table}[h]
\centering
\caption{Ablation study on post-processing strategies during inference. \textit{Keyframe Bidirectional} significantly outperforms the baseline, highlighting the model's strong keyframe localization ability.}
\resizebox{0.48\textwidth}{!}{
  \begin{tabular}{lcc}
  \toprule
  \textbf{Strategy} & \textbf{MeViS} ($val_u$) & \textbf{Long-RVOS} \\
  \midrule
  Sa2VA-type First N  & 59.95 & 47.22 \\
  + Keyframe Initialization  & 61.54 & 48.15 \\
  + [SEG] Injected Propagation & \textbf{63.07} & \textbf{49.55} \\
  \bottomrule
  \end{tabular}
}
\label{tab:ablation_postprocess}
\end{table}

\subsection{Effectiveness of Joint Segmentation Loss}
\label{app:seg_loss}
\begin{figure}[h]
    \centering
    \includegraphics[width=0.4\textwidth]{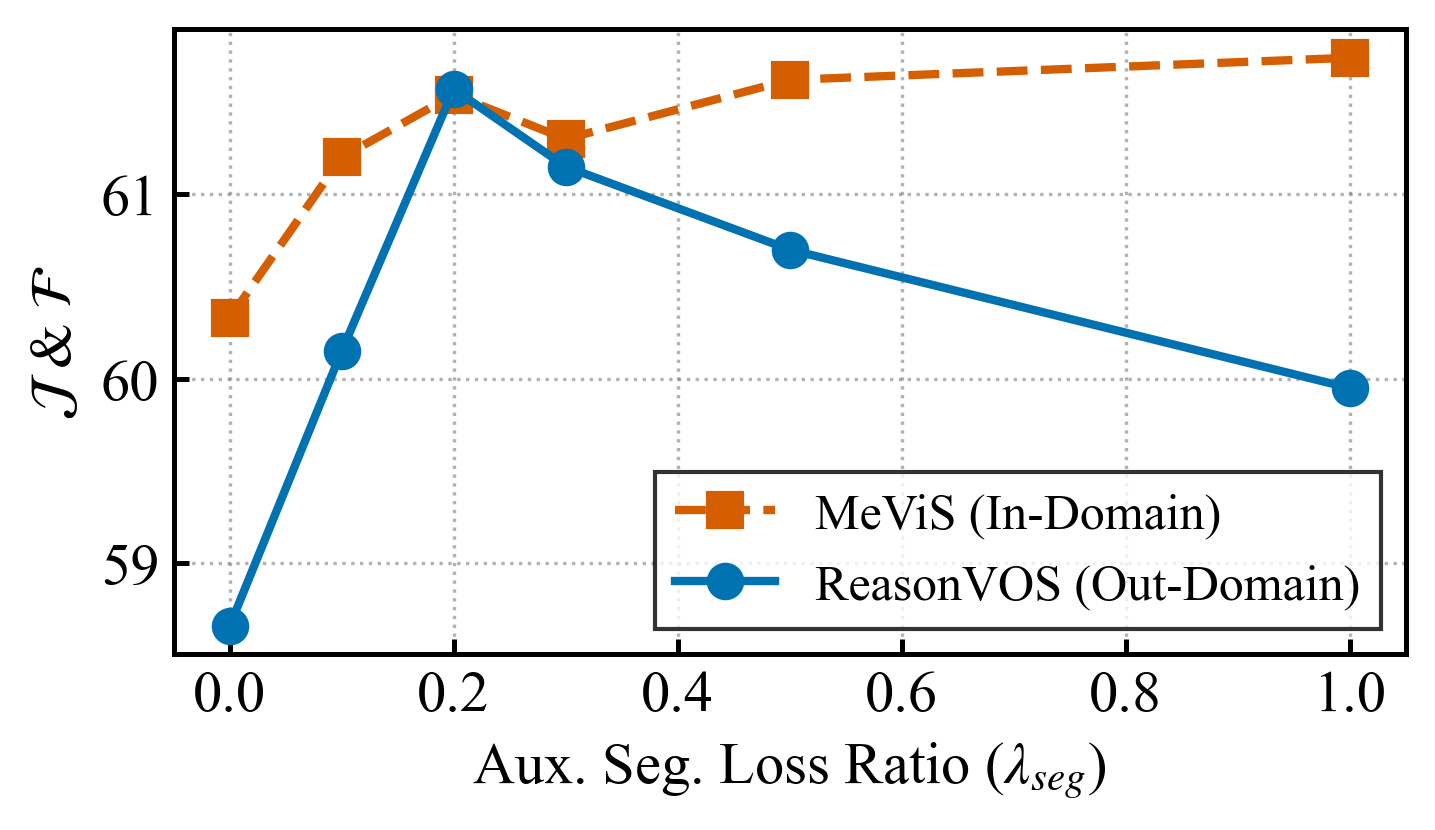} 
    \caption{Ablation study on auxiliary segmentation loss ($\lambda_{seg}$).}
    \label{fig:aux_seg_loss}
\end{figure}

To validate the effectiveness of the auxiliary segmentation loss proposed in Sec.~\ref{subsec:train_objective}, we conduct ablation studies by varying its weight ratio to modulate its influence during training. As depicted in Fig.~\ref{fig:aux_seg_loss}, increasing $\lambda_{seg}$ yields consistent performance gains on the in-domain MeVIS dataset. Conversely, on the out-of-domain ReasonVOS dataset, we observe an initial improvement followed by a decline. Notably, this tendency becomes more pronounced as the number of training epochs increases. These results suggest that while the auxiliary loss enhances adaptation to the training distribution, an excessive weight induces overfitting, thereby compromising generalization capabilities. Consequently, we adopt $\lambda_{seg}=0.2$ to achieve an optimal balance between in-domain performance and generalizability.

\begin{table}[h]
\centering
\caption{\textbf{Ablation study on reward design.} $\mathcal{R}_p, \mathcal{R}_{km}$, and $\mathcal{R}_t$ denote Progressive Mask Improvement, Keyframe Mask Quality, and Temporal Accuracy rewards, respectively. }
\label{tab:ablation_reward}
\resizebox{0.48\textwidth}{!}{
  \begin{tabular}{l|ccc}
  \toprule
  Model & MeViS ($val_u$) & Ref-SAV & Avg. Rounds \\
  \midrule
  Step-wise Reward & 57.88 & 50.18 & {1.63} \\
  Episodic Reward  & \textbf{61.19} & \textbf{52.91} & \textbf{2.13} \\
  \midrule
  w/o $\mathcal{R}_{p}$ & 60.37 {\color{gray}\scriptsize(-0.82)} & 52.62 {\color{gray}\scriptsize(-0.29)} & 2.46 {\color{gray}\scriptsize($\uparrow$15.5\%)} \\
  w/o $\mathcal{R}_{km}$ & 60.19 {\color{gray}\scriptsize(-1.00)} & 50.99 {\color{gray}\scriptsize(-1.92)} & 2.36 {\color{gray}\scriptsize($\uparrow$10.8\%)} \\
  w/o $\mathcal{R}_{t}$ & 58.70 {\color{gray}\scriptsize(-2.49)} & 51.11 {\color{gray}\scriptsize(-1.80)} & 2.55 {\color{gray}\scriptsize($\uparrow$19.7\%)} \\
  \bottomrule
  \end{tabular}
}
\end{table}

\subsection{Effectiveness of Reward Design}
\label{app:reward_ablation}
As shown in Table~\ref{tab:ablation_reward}, we investigate the impact of distinct reward components on both segmentation accuracy and interaction efficiency. The \textit{Progressive Mask Improvement reward ($\mathcal{R}_p$)} is critical for regulating reasoning length; removing $\mathcal{R}_p$ results in redundant exploration, increasing the average rounds from 2.13 to 2.46. This confirms that penalizing stagnant mask quality effectively suppresses unnecessary dialogue turns. Furthermore, the \textit{Keyframe Mask Quality reward ($\mathcal{R}_{km}$)} and \textit{Temporal Accuracy reward ($\mathcal{R}_t$)} drive the model toward precise spatiotemporal grounding. Notably, the omission of $\mathcal{R}_t$ leads to the most significant performance drop on MeViS ($61.19\% \rightarrow 58.70\%$), underscoring its role in ensuring the model identifies the correct temporal interval before attempting dense segmentation.

We further explore the impact of replacing episodic rewards with step-wise rewards during training. In this setting, the reward is calculated at each step based on varying maximum interaction horizons (e.g., for a 4-turn limit, rewards are computed for horizons $T \in \{1, 2, 3, 4\}$). Our empirical analysis reveals that this step-wise approach leads to significant \textit{reward hacking} and eventual reward collapse. Specifically, the policy tends to converge toward predicting with minimal interaction turns, failing to exploit the advantages of multi-turn reasoning. This phenomenon occurs because the model prioritizes maximizing immediate rewards within fewer steps, thereby diminishing its focus on the incremental gains between turns. Consequently, the model loses the capacity for effective multi-round refinement, leading to a degradation in overall performance.

% =========================================================================
% Section E: Qualitative Results
% =========================================================================
\section{Additional Qualitative Results}
\label{app:qualitative}

\subsection{Multi-turn Reasoning Visualizations}
\label{app:vis_multiturn}

To intuitively illustrate the multi-turn reasoning capabilities of VideoSEG-O3, we provide visualizations of representative conversational cases. We explicitly present the complete progression, including system prompts, per-round user instructions, visual inputs, chain-of-thought reasoning processes, and the final predicted masks.

\begin{figure*}[h]
    \centering
    \includegraphics[width=1.0\textwidth]{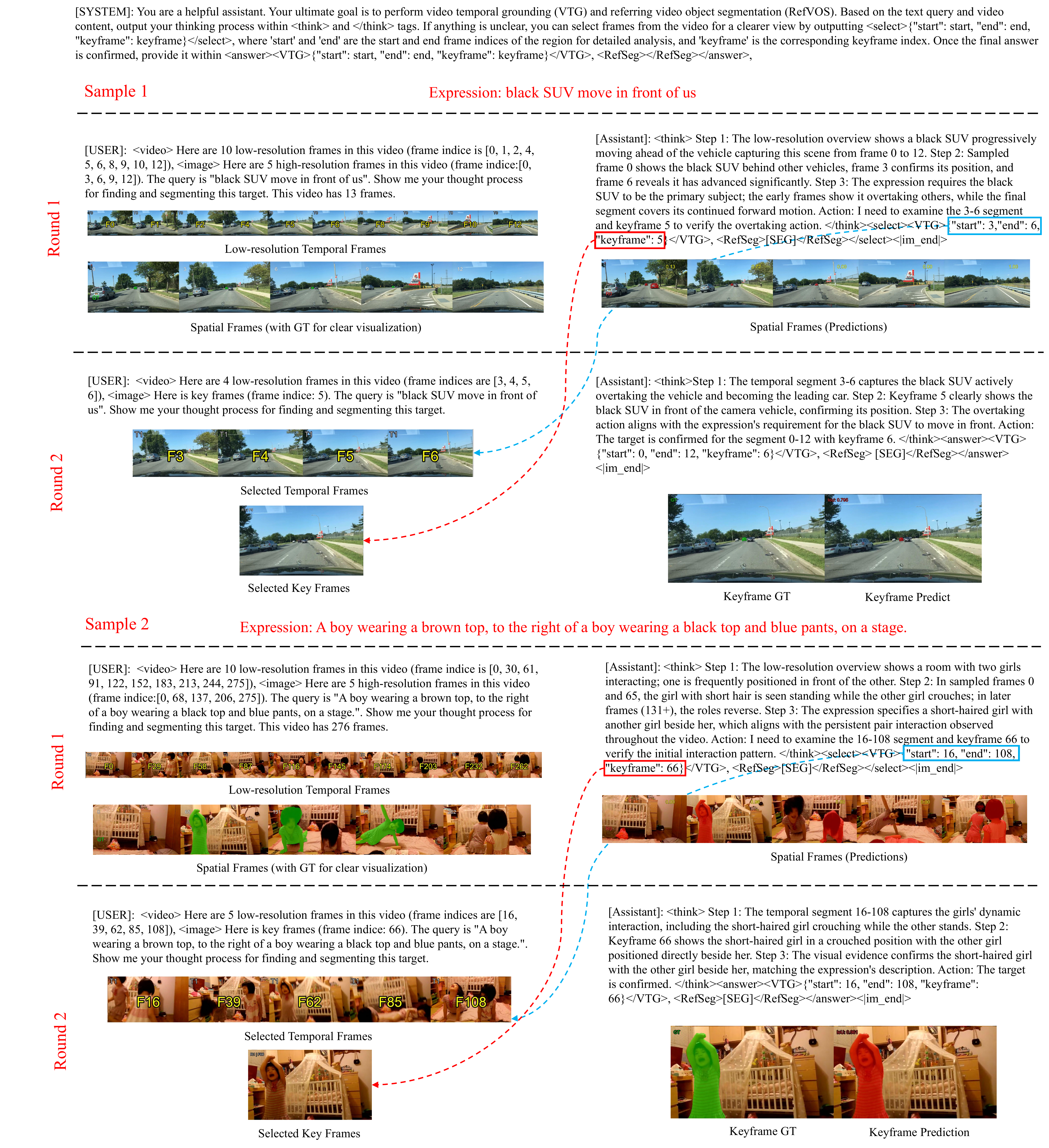} 
    \caption{Visualization of a \textbf{two-round} interactive reasoning and segmentation process.}
    \label{fig:sup_vis_2round}
\end{figure*}

\noindent\textbf{Two-round Dialogues.}
As shown in Fig.~\ref{fig:sup_vis_2round}, we first demonstrate the model's performance in two-round interaction scenarios. In these cases, VideoSEG-O3 effectively utilizes immediate user feedback to correct initial ambiguities. By integrating historical reasoning with new instructions, the model refines its understanding of the scene, allowing it to swiftly adjust its focus and achieve accurate object grounding after a single observation step.

\begin{figure*}[h]
    \centering
    \includegraphics[width=1.0\textwidth]{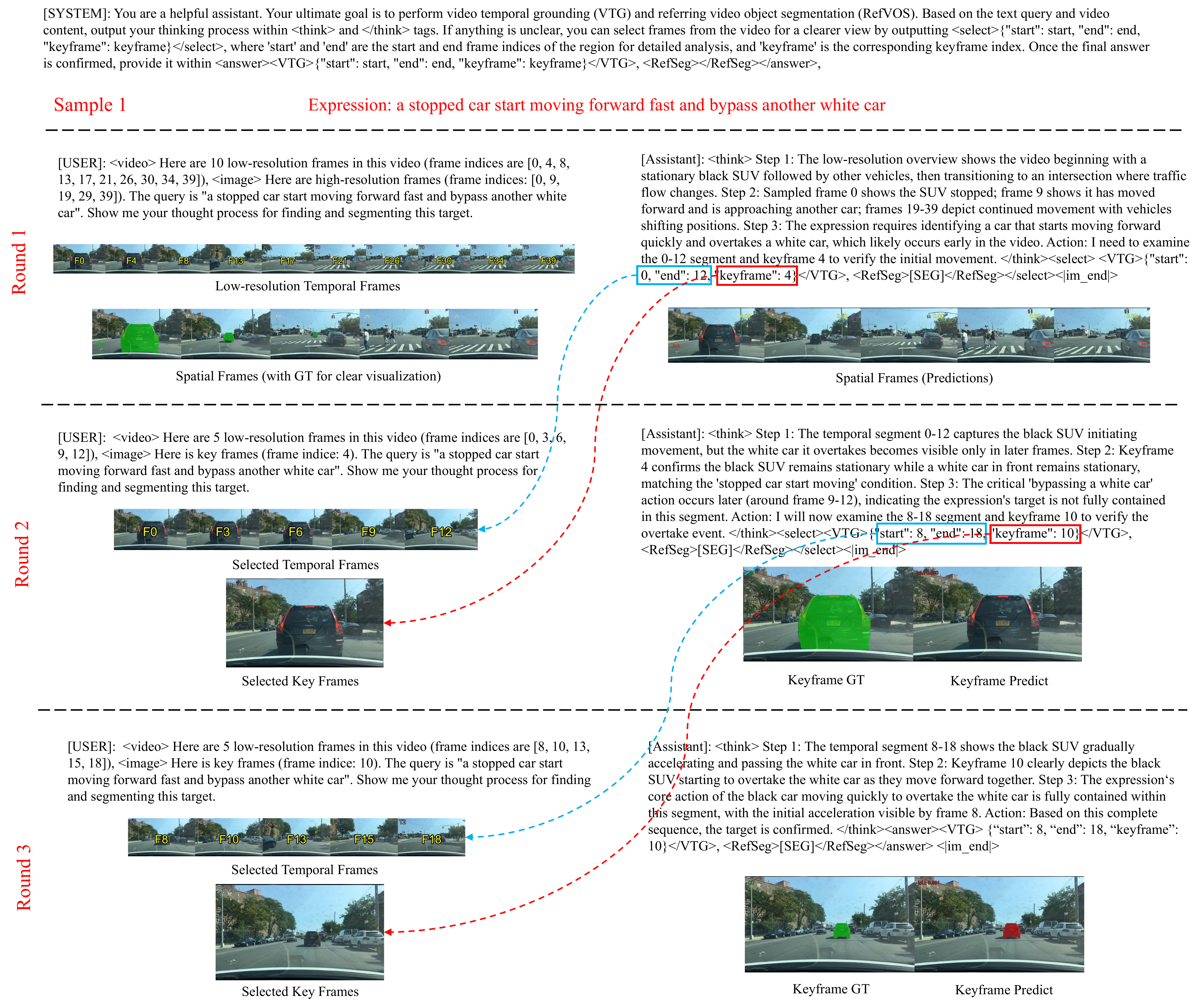} 
    \caption{Visualization of a \textbf{three-round} iterative refinement process for complex target localization (Sample 1).}
    \label{fig:sup_vis_3round_1}
\end{figure*}

\begin{figure*}[h]
    \centering
    \includegraphics[width=1.0\textwidth]{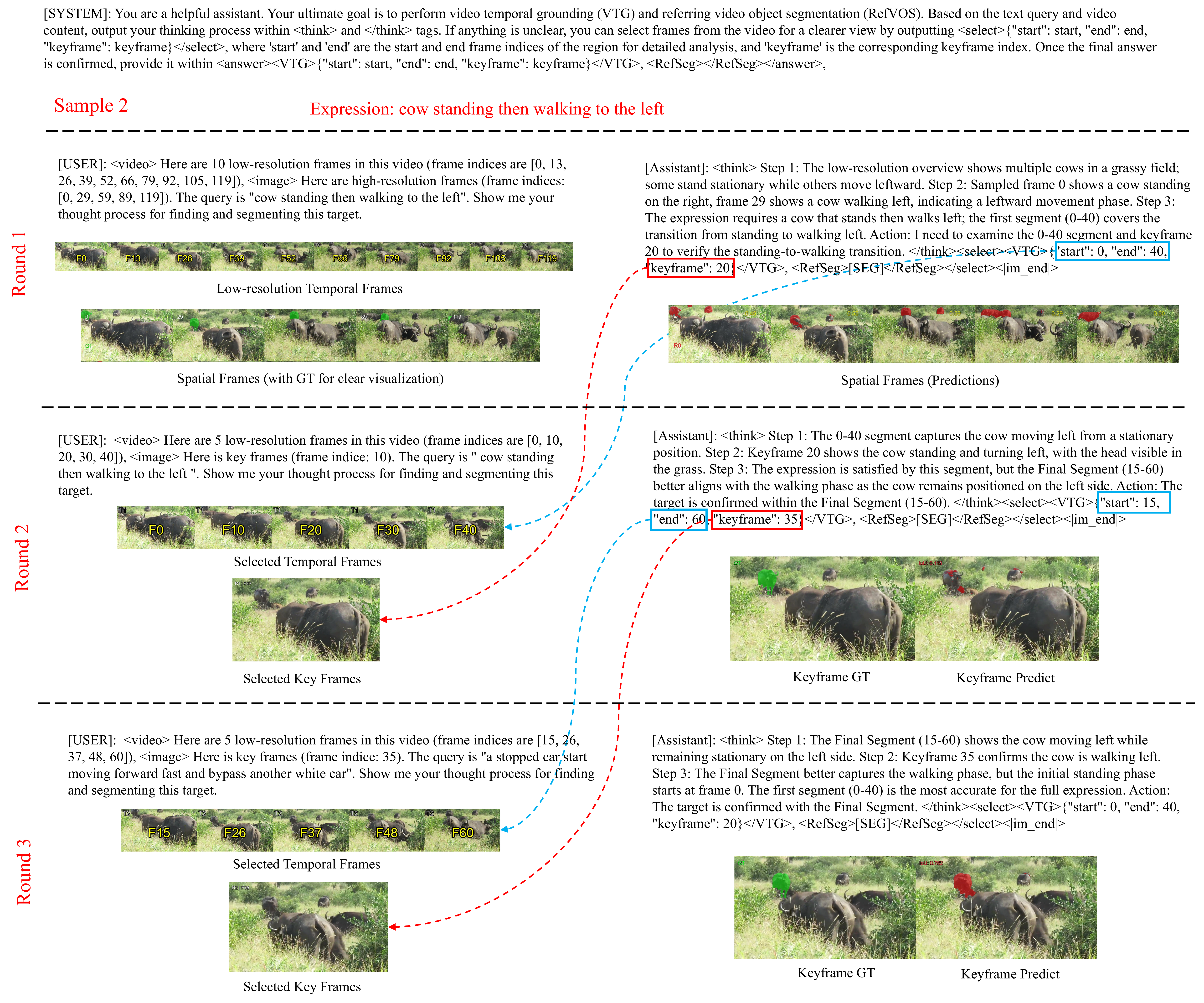} 
    \caption{Visualization of a \textbf{three-round} iterative refinement process for complex target localization (Sample 2).}
    \label{fig:sup_vis_3round_2}
\end{figure*}

\noindent\textbf{Three-round Dialogues.}
Fig.~\ref{fig:sup_vis_3round_1} and Fig.~\ref{fig:sup_vis_3round_2} further showcase the model's proficiency in handling intricate target localization tasks through a three-round iterative process. These visualizations reveal a clear coarse-to-fine refinement trajectory, where the model progressively narrows down the target by synthesizing cumulative visual evidence and multi-step feedback. From an information-theoretic perspective, this progressive interaction proves more efficient than processing exhaustive video content in a single pass, offering a robust paradigm for precise segmentation in complex dynamic environments.

% Fig.~\ref{fig:sup_vis_3round_1} further showcases the model's proficiency in handling intricate target localization tasks through a three-round iterative process. These visualizations reveal a clear coarse-to-fine refinement trajectory, where the model progressively narrows down the target by synthesizing cumulative visual evidence and multi-step feedback. From an information-theoretic perspective, this progressive interaction proves more efficient than processing exhaustive video content in a single pass, offering a robust paradigm for precise segmentation in complex dynamic environments.

\end{document}